\PassOptionsToPackage{nosubfloats}{jmlrutils}

\documentclass{template}
\usepackage{graphicx}
\usepackage{booktabs,array}
\usepackage{tikz}
\usetikzlibrary{arrows.meta,decorations.markings,calc}
\usepackage{cuted}
\usepackage{caption}
\usepackage{subcaption}
\usepackage{wrapfig}
\usepackage{ragged2e}

\usepackage{etoc}

\usepackage[english]{babel}

\newcommand{\trace}{\operatorname{tr}}
\newcommand{\ntrace}{\bar\trace\,}

\newcommand{\df}{\operatorname{df}}
\newcommand{\qed}{\hfill$\Box$}

\usepackage{amsmath}
\usepackage{graphicx}
\usepackage{listings}
\usepackage{xcolor}

\definecolor{codegreen}{rgb}{0,0.6,0}
\definecolor{codegray}{rgb}{0.5,0.5,0.5}
\definecolor{codepurple}{rgb}{0.58,0,0.82}
\definecolor{backcolour}{rgb}{0.95,0.95,0.92}

\lstdefinestyle{mystyle}{
    backgroundcolor=\color{backcolour},   
    commentstyle=\color{codegreen},
    keywordstyle=\color{magenta},
    numberstyle=\tiny\color{codegray},
    stringstyle=\color{codepurple},
    basicstyle=\ttfamily\footnotesize,
    breakatwhitespace=false,         
    breaklines=true,                 
    captionpos=b,                    
    keepspaces=true,                 
    numbers=left,                    
    numbersep=5pt,                  
    showspaces=false,                
    showstringspaces=false,
    showtabs=false,                  
    tabsize=2
}

\AtBeginDocument{%
  \setlength{\parindent}{0pt}
  \setlength{\parskip}{0.5\baselineskip}
}

\AtBeginDocument{%

  \setcounter{topnumber}{4}
  \setcounter{bottomnumber}{2}
  \setcounter{totalnumber}{6}
}

\makeatletter
\AtBeginDocument{%
  \jmlrproceedings{}{}%
  \jmlrvolume{}%
  \jmlrissue{}%
  \jmlryear{}%
  \jmlrpages{}%
  \jmlrworkshop{}%
  \jmlrsubmitted{}%
  \jmlrpublished{}%
  \editor{}%

  \let\@titlefoot\@empty
  \let\ps@jmlrtps\ps@plain
}
\makeatother

\title[New Multiple Descent Phenomena Induced by Heterogeneous Data]{New Multiple Descent Phenomena Induced by Heterogeneous Data}

\author{%
Mark-Anthony Moisescu-Pareja$^{1,2}$\\
Elvis Dohmatob$^{1,2}$\\
    \addr{$^1$Concordia University}\\
  \addr{$^2$Mila -- Quebec AI Institute}
}

\begin{document}
\etocdepthtag.toc{main}

\maketitle

\begin{abstract}
Modern machine learning models are often trained on large datasets assembled from many different sources, whose statistical properties can differ substantially. Understanding how this heterogeneity affects generalization performance is therefore important for explaining how model performance changes as model size increases. In this work, we study this question in linear random-feature models and show that mixtures of datasets with different covariance structures can give rise to a range of multiple-descent phenomena. We derive exact asymptotic expressions for the test error of each data source in the high-dimensional regime, and show how heterogeneity in covariance scale and spectral structure can introduce additional, group-dependent interpolation thresholds as model size increases. These phenomena arise even with noiseless labels and linear activations, showing that neither label noise nor activation nonlinearity is necessary for multiple-descent. Our results therefore provide a different perspective from classical explanations of multiple descent. More broadly, our results have potential implications for algorithmic fairness by showing that model scaling can affect different data groups unevenly, amplifying disparities in their generalization errors. We validate our theoretical results on synthetic Gaussian mixtures and real-world datasets, including MNIST and CIFAR-10. 
\end{abstract}

\section{Introduction}
\label{sec:intro}
Many learning systems are trained on data pooled from multiple sources, such as hospitals, devices, or user populations. These sources may have different statistical properties even when they share the same prediction task. Pooling their data therefore raises questions beyond those encountered in a homogeneous setting: how do differences between sources interact with model size, and how are the resulting benefits and errors distributed across sources? Understanding generalization in this setting requires accounting for both the composition of the training mixture and the statistical structure of its components.

The relationship between model size and generalization can itself be highly non-monotonic. Double- and triple-descent phenomena show that increasing model size can lead to successive improvements and deteriorations in prediction performance \citep{belkin2019reconciling, d2020triple, adlam2020neural}. Existing analyses have identified mechanisms involving label noise and nonlinear feature maps, in the setting of homogeneous datasets.

\paragraph{Contributions.}

In this work, we investigate covariance heterogeneity across training data sources as a mechanism for genuine multiple-descent. Focusing on linear random-feature models consisting of a frozen linear hidden layer and a trained linear readout, we show that heterogeneous data mixtures can induce multiple-descent behavior even with a shared ground-truth predictor and noiseless labels. This setup isolates the role of covariance structure in shaping generalization as model size varies. Our main contributions are as follows:

\begin{itemize}
\item\emph{Exact test error curves.}
We develop a high-dimensional theory of generalization for linear random-feature models trained on heterogeneous data mixtures, deriving asymptotically exact expressions for the test error on each source. These formulas characterize how model size and mixture composition jointly shape source-specific generalization (Section~\ref{sec:bias-variance}), extending existing analyses of homogeneous data \citep{Hastie2019Surprises,cui2021generalization,bach2024high}.
\item\emph{New multiple-descent phenomena.}
We expand the classical picture of multiple descent \citep{adlam2020neural,d2020triple} by showing that covariance heterogeneity alone can induce this behavior, even with noiseless labels and linear activations. Our analysis identifies conditions under which these phenomena emerge and characterizes the resulting source-specific test error curves (Section~\ref{sec:triple-descent}). We also discuss their implications for algorithmic fairness: these phenomena can potentially amplify group-wise error disparities, highlighting the importance of per-group evaluation when assessing large models.
\item\emph{Empirical validation.} In Section~\ref{sec:exp}, we validate our theoretical predictions through experiments on controlled synthetic Gaussian mixtures and on MNIST \citep{deng2012mnist} and CIFAR-10 \citep{krizhevsky2009cifar}. The experiments closely match the predicted behavior and demonstrate that our new multiple-descent phenomena persist robustly beyond the Gaussian setting.
\end{itemize}

A detailed overview of the literature is presented in Section \ref{sec:related-work}.

\section{Setup}

\subsection{Heterogeneous Gaussian Groups}
Consider a collection of $K \geq 1$ independent groups, each characterized by its own data distribution. Mathematically, for $k \in [K]$, we define the k-th group $\mathcal{D}_k = \{(x_i, y_i) \in \mathbb{R}^{d} \times \mathbb{R}: i = 1,...,n_k\}$ as follows:

The input features for the $k$-th group are sampled from a multivariate Gaussian distribution: 
\begin{eqnarray}
    \textbf{(Inputs) }x_i \sim \mathcal{N}(0, \Sigma_k),
\label{eq:gaussinputs}
\end{eqnarray}
where $\Sigma_k \in \mathbb{R}^{d \times d}$ is a positive-definite matrix that captures the covariance structure of the group. The associated labels for the group are generated as: 
\begin{eqnarray}
   \textbf{(Labels) } y_i = f_k^*(x_i) + z_k,
\label{eq:gausslabels}
\end{eqnarray}
where $f_k^*(x_i) = x_i^\top w_*^{(k)}$ is the teacher model, $w_*^{(k)} \in \mathbb{R}^d$ are the ground-truth weights, and $z_k \sim \mathcal{N}(0, \sigma_k^2)$ is independent per-sample label noise for the $k$-th group. The full training set is then formed by combining the $K$ data sources, giving the heterogeneous group $\mathcal{D} = \bigcup_{k=1}^K \mathcal D_k$ with $n=\sum_{k=1}^K n_k$ training samples.

\begin{remark}
For conciseness, most theoretical results will consider the case of two groups (i.e. $K = 2$), of sizes $n_1$ and $n_2$, for a heterogeneous group $\mathcal D=\mathcal D_1 \cup \mathcal D_2$ of total size $n = n_1+n_2$. However, all results can be easily extended to any finite number of groups. 
\end{remark}

\begin{remark}
Throughout this work, we focus on the case of shared ground-truth weights vector $w_*^{(k)}=w_*$ for all $k \in [K]$, which is already sufficient to produce the multiple descent phenomena studied here. Our results are not dependent on this assumption and can be easily extended to the more general setting in which each group has its own ground-truth weights $w_*^{(k)}$.
\end{remark}

\subsection{Linear Random-Features Model}
For the student model, we consider a one-layer random-features (RF) neural network
\begin{eqnarray}
    f_\textrm{RF}(x) = \hat v^\top Wx,
\label{eq:linearrf}
\end{eqnarray}
where the first-layer weights $W$ form a random $m \times d$ matrix with i.i.d. entries drawn from $\mathcal N(0,1/d)$, a scaling analogous to the Xavier initialization \citep{glorot2010understanding}, which is widely used in deep learning practice. Random-feature models were first introduced by \citet{rahimi2007random} as a computationally efficient approximation to kernel methods, replacing an implicit kernel representation with a finite-dimensional randomized feature map followed by a linear predictor. They have since become a useful theoretical model for studying high-dimensional generalization and over-parameterization \citep{mei2022generalization,bach2024high}. In our case, we consider linear activation, so that we can isolate the effect of heterogeneity in the training data without introducing additional nonlinear structure through the feature map. 

The output weights vector $\hat v \in \mathbb R^m$ is obtained via ridge regression on the heterogeneous training group $\mathcal D = \mathcal D_1 \cup \mathcal D_2$ of total size $n=n_1+n_2$:
\begin{eqnarray}
    \hat v = \arg\min_{v \in \mathbb R^m}\frac{1}{n}\sum_{(x_i,y_i) \in \mathcal D}(f(x_i) - y_i)^2 + \lambda\|v\|^2.
\end{eqnarray}
Here,  the scalar $\lambda > 0$ controls the amount of regularization; we shall be particularly interested in the unregularized/so-called ridgeless limit $\lambda\to 0^+$.

We are mainly interested in analyzing the dynamics of the quenched test errors 
\begin{eqnarray}
\begin{split}
\mathcal{E}_k &= \mathbb E_{\mathcal D}\left[\mathbb{E}_{\textbf{x} \sim \mathcal{N}(0,\Sigma_k)}\left[\left(f_\textrm{RF}(\textbf{x}) - f_k^*(\textbf{x}) \right)^2 \right]\right],
\end{split}
\label{eq:Ek}
\end{eqnarray}
of the RF student model in the high-dimensional proportionate scaling limits:
\begin{equation}
\boxed{
\begin{gathered}
d, m, n, n_1, n_2 \to \infty,\,\, d/n \to \phi,\,\, m/n \to \psi,\,\, m/d \to \gamma,\,\, n_1/n \to p_1,\,\, n_2/n \to p_2,
\end{gathered}
}
\label{eq:proportionate}
\end{equation}
where $\phi, \psi,\gamma \in (0,\infty)$ and $p_1,p_2 \in (0,1)$ such that $p_1+p_2=1$ and $\gamma=\psi/\phi$. This scaling regime captures how the relative sizes of the input dimension $d$, model width $m$, total sample size $n$, and group proportions $p_k$ influence the generalization behavior of the  model.

\subsection{Motivation for the Setup}
The setup of our work isolates how covariance heterogeneity shapes generalization as model size varies. Linear random-feature models allow us to study multiple descent even with a shared ground-truth predictor and noiseless labels, while source-specific test errors reveal its effects across groups. Two-source mixtures already exhibit this mechanism; arbitrary $K$ enables a systematic study of richer mixtures. Related frameworks address bias amplification \citep{arjun2025} and model collapse \citep{dohmatob2025strong}, whereas our focus is the emergence of multiple-descent curves. The single-source case provides a reference point connecting our analysis to double descent \citep{Hastie2019Surprises, belkin2019reconciling,bach2024high} and covariate shift \citep{tripuraneni2021covariate}.

\section{Theoretical Results}
Let $(\delta,\rho_1,\rho_2)$ be defined as the unique positive triplet which solves the following system of fixed-point equations:
\begin{equation}
\boxed{
\begin{gathered}
\psi \lambda = (\psi - 1)\delta + q\delta^2,
\quad
q := \frac{p_1}{\rho_1} + \frac{p_2}{\rho_2},\\[1pt]
\text{with }\rho_j := \delta + \psi\mu_j,
\quad
\mu_j := \phi\ntrace\Sigma_j K^{-1},
\quad
K := \phi I_d + \psi K_0,
\quad
K_0 := \frac{p_1}{\rho_1}\Sigma_1
+ \frac{p_2}{\rho_2}\Sigma_2.
\end{gathered}
}
\label{eq:delta}
\end{equation}
Here, $\ntrace$ denotes the normalized trace operator. The scaling constants $p_1,p_2,\phi,\psi$ are as in the asympotic limits \eqref{eq:proportionate}.

\subsection{Asymptotic Bias-Variance Decomposition}
\label{sec:bias-variance}
We first derive a high-dimensional bias-variance decomposition of the group-wise test error.
\begin{theorem}
Let $a_1$, $a_2$, $A$, $D_{12}$, $D_{21}$, $E_1$, $E_2$, $\tilde E_0$, $\tilde E_1$, $\tilde E_2$, $u$, $v$, $w$, $r$, $z_1$, $z_2$ be non-negative scalars (defined explicitly in Appendix \ref{sec:constants}), that only depend on the scaling proportions $\psi,\phi,p_1,p_2$ in \eqref{eq:proportionate}, the covariance matrices $\Sigma_1$, $\Sigma_2$, and the quantities defined in \eqref{eq:delta}. In the limit \eqref{eq:proportionate}, the test error defined in \eqref{eq:Ek} admits a bias-variance decomposition
\begin{eqnarray}
\mathcal E_k \simeq B_k + V_k,
\label{eq:bias-variance}
\end{eqnarray}
where the bias terms $B_1,B_2 \ge 0$ and variance terms $V_1,B_2 \ge 0$ are given by:
\begin{equation}
\boxed{
\begin{gathered}
B_1 := p_1\psi x a + p_2\psi y b + \tilde z_0,
\quad
V_1 := \sum_{s=1}^2 p_s \sigma_s^2
\frac{\det U^{(s)}_{1 \leftarrow V^{(s)}}}{\det U^{(s)}},
\\[6pt]
\text{with}\quad
\begin{pmatrix} a\\b \end{pmatrix}
=
\begin{pmatrix}
1-p_1\psi u & -p_2\psi v\\
-p_1\psi w & 1-p_2\psi r
\end{pmatrix}^{-1}
\begin{pmatrix} z_1\\z_2 \end{pmatrix},
\\[6pt]
V^{(1)} := \begin{pmatrix}
\tilde E_1\\
\tilde E_2\\
\phi\tilde E_0
\end{pmatrix},
\quad
V^{(2)} := \begin{pmatrix}
\tilde E_2\\
\tilde E_1\\
\phi\tilde E_0
\end{pmatrix},
\quad
U^{(1)} := \begin{pmatrix}
a_1 & -p_2 D_{12} & -E_1/\gamma\\
-p_1 D_{12} & a_2 & -E_2/\gamma\\
-p_1\phi E_1 & -p_2\phi E_2 & 1-\psi A
\end{pmatrix}.
\end{gathered}
}
\label{eq:bias_variance}
\end{equation}

and $B_2$, $V_2$ are defined analogously with $1$ and $2$ swapped everywhere.  In the above display, $U_{j \leftarrow V}$ is notation for the matrix obtained from $U$ by replacing its $j$th column with the vector $V$, and $V^{(2)}$ (resp. $U^{(2)}$) is obtained from $U^{(1)}$ by swapping the subscripts $1$ and $2$ in all symbols appearing in the expression for $U^{(1)}$.
 \label{thm:main}
\end{theorem}

\paragraph{Sketch of proof of Theorem \ref{thm:main}.} 
It is easy to see that $\hat v$ has the explicit representation
\begin{eqnarray}
    \hat v = (\Phi^\top \Phi/n + \lambda I)^{-1}\Phi^\top Y/n,
    \label{eq:ridge}
\end{eqnarray}
where $\Phi=XW^\top \in \mathbb R^{n \times m}$ is the RF matrix for the entire dataset, $X = \begin{pmatrix}
    X_1\\X_2
\end{pmatrix} \in \mathbb R^{n \times d}$ is the corresponding design matrix, $Y=\begin{pmatrix}
    Y_1\\ Y_2
\end{pmatrix} \in \mathbb R^n$ is the response vector, $X_k=(x_i) \in \mathbb R^{n_k \times d}$ is the design matrix for the $k$th group, and finally $Y_k=(y_i) \in \mathbb R^{n_k}$ is the corresponding response vector.

Now, define $\hat w:=W^\top \hat v \in \mathbb R^d$, where $\hat v=R\Phi^\top Y/n \in \mathbb R^m$ is as defined in \eqref{eq:ridge} and let
$$
R:=(\Phi^\top \Phi/n + \lambda I)^{-1} \in \mathbb R^{m \times m}
$$
define the  \emph{resolvent} of $\Phi$.
The Group-$k$ test error of the model defined in \eqref{eq:Ek} writes
\begin{align*}
    \mathcal E_k 
    &=  \mathbb E_{X,Y,W}\,[(w_* - \hat w)^\top \Sigma_k(w_*-\hat w)],
\end{align*}
for a random test point $x \sim \mathcal N(0,\Sigma_k)$, independent of the training data $\mathcal D=(X,Y)$ and weights matrix $W$.
Define $S := S_1 + S_2$, where $S_k := X_k^\top X_k/n$ is the empirical covariance matrix for the $k$th group. Setting $T_k := WS_kW^\top$, one computes
\begin{align*}
    w_* - \hat w &= w_* - W^\top \hat{v} =  w_* - W^\top R \Phi^\top Y / n 
    = w_* - W^\top RW \left(Sw_* + \frac{1}{n} X_1^\top Z_1 + \frac{1}{n} X_2^\top Z_2 \right)
\end{align*}
where $Z_k = Y_k - X_kw_*$ is the noise vector for $k$th group.

We can then write test error for Group $k$ as $\mathcal E_k = B_k + V_k$, where the bias $B_k$ and variance terms $V_k$ are given by:
\begin{align*}
B_k& = \lambda^2\,\mathbb E\,[w_*^\top H^\top \Sigma_k H w_*]=\lambda^2 \mathbb E \frac1d \trace \Theta H^\top \Sigma_k H,\\
    V_k &= \frac{1}{n^2}\sum_{s=1}^2 \sigma_s^2\mathbb E\,\trace X_s W^\top R W \Sigma_k W^\top R W X_s^\top = \sum_{s=1}^2 \sigma_s^2\,\mathbb E \frac{1}{n} \trace T_s R W \Sigma_k W^\top R,
\end{align*}
with $\Theta:=d w_* w_*^\top \in \mathbb R^{d \times d}$, $H:=(W^\top WS + \lambda I_d)^{-1} \in \mathbb R^{d \times d}$, and the expectations are over the design matrix combined $X \in \mathbb R^{n \times d}$ and the random first layer weights matrix $W \in \mathbb R^{m \times d}$.
The tools of Free Probability Theory \citep{mingo2017free} can then be used to analyze the above traces in the limit \eqref{eq:proportionate},  obtaining the analytic expressions in the theorem. The full details are provided in Appendix \ref{sec:bvproofs}.

\subsection{Special Case of Homogeneous Groups: Recovering the Classical Picture}
We now consider the case of a single group, corresponding to the limit $p_2 \to 0^+$ (i.e. $p_1 \to 1$). This limit represents the homogeneous case in which the training data effectively come from a single population. To simplify notation, write $\Sigma_1=\Sigma$ and $\sigma_1=\sigma$. Let $\kappa:=\rho_1/\gamma$, where $\rho_1$ is defined in \eqref{eq:delta}. It can be shown that $\kappa$ is the unique positive solution to the following fixed-point equation:
\begin{eqnarray}
    \lambda = \kappa(\gamma - I)(1-\phi I),\text{ with }I := \ntrace\Sigma(\Sigma + \kappa I_d)^{-1}.
\end{eqnarray}
We further define the following scalars
\begin{eqnarray}
\begin{split}
        J &:= \ntrace\Sigma(\Sigma + \kappa I_d)^{-2},\quad \tilde J := w_*^\top \Sigma(\Sigma + \kappa I_d)^{-2}w_*,\\
        D &:= \ntrace\Sigma_1^2(\Sigma + \kappa I_d)^{-2},\quad  \tilde D := w_*^\top \Sigma_1^2(\Sigma + \kappa I_d)^{-2}w_*.
\end{split}
\end{eqnarray}
Note that these quantities satisfy several useful identities. For example, it is an elementary result that $I - D = \kappa J$. We now provide explicit expressions for the bias and variance terms in the homogeneous setting through the following corollary: 
\begin{corollary}
\label{corr:single}
Consider the setting of a single group (i.e., $p_2 \to 0^+$). In the limit \eqref{eq:proportionate}, we have the bias-variance decomposition $\mathcal E_1 \simeq B_1 + V_1$, where
\begin{eqnarray}
    \begin{split}
       B_1 &= \kappa^2 \frac{(1-D/\gamma)\tilde J + J \tilde D/\gamma}{(1-D/\gamma)(1-\phi D) - \phi \kappa^2 J^2/\gamma},\\
            V_1 &= \sigma^2 \phi\frac{(\gamma - I)I+ (2I - \gamma)\kappa J}{(\gamma - I)(1 - \phi I) + (1 + \psi - 2\phi I)\kappa J}.
    \end{split}
\end{eqnarray}
\end{corollary}
Even in this special case of a single group, the above result already generalizes the main results from \citep{bach2024high} which focuses on the ridgeless limit $\lambda \to 0^+$.

\subsection{Triple-Descent without Non-Linearities}
\label{sec:triple-descent}
We now show that triple-descent behavior, previously associated with the combined effects of model scaling, label noise, and the geometry induced by nonlinear activation functions \citep{d2020triple, adlam2020neural}, can also arise in a linear random-features model as a consequence of heterogeneity in the input covariance structure alone.

\begin{figure*}[!htp]
    \centering
    \includegraphics[width=1.0\linewidth]{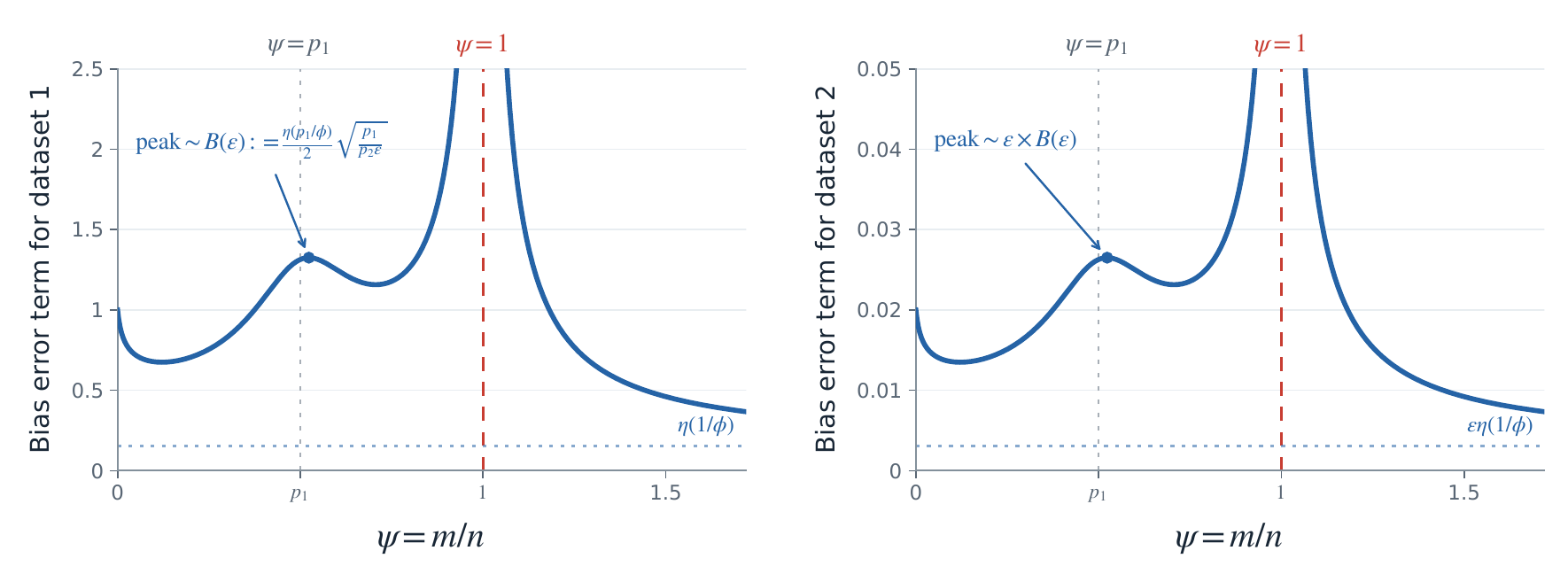}
    \vspace{-.5cm}
\caption{
\textbf{Triple descent in a linear RF model.}
Cartoon bias curves for a heterogeneous mixture of $n_1$ samples with power-law covariance
$\Sigma_1=\Sigma=\operatorname{diag}(1,1/2,\ldots,1/d)$
and $n_2$ samples with rescaled covariance $\Sigma_2=\varepsilon\Sigma$, where $\varepsilon\ll1$.
The group-wise bias terms $B_1$ (\textbf{left}) and $B_2$ (\textbf{right}) are plotted against $\psi=m/n$, with network width $m$ and total sample size $n=n_1+n_2$.
For small $\varepsilon$, the  peaks near $\psi=p_1:=n_1/n$ ($m=n_1$) have respective heights asymptotic to
$B(\varepsilon):=\frac{\eta(p_1/\phi)}{2}\sqrt{\frac{p_1}{p_2\varepsilon}}$
and $\varepsilon B(\varepsilon)$, as predicted by Corollary~\ref{corr:Zipf}.
Here, $\phi=d/n>1$, $d$ is the input dimension, and $\eta$ is the explicit function defined in that corollary.
Both curves diverge at the classical interpolation threshold $\psi=1$ ($m=n$).
Horizontal dotted lines mark the large-width limits $\eta(1/\phi)$ and $\varepsilon\eta(1/\phi)$.
}
\label{fig:triple-cartoon}
\end{figure*}

\paragraph{Multiple-descent from heterogeneous covariance scales.} Consider the case in which the covariance matrices of two groups are proportional:
\begin{eqnarray}
\label{eq:cov-proportional}
    \Sigma_1 = \Sigma,\quad \Sigma_2 = \varepsilon \Sigma,
\end{eqnarray}
where $0 < \varepsilon \ll 1$ controls the separation in scale between the two groups. In this setting, Group 2 can be viewed as a rescaled version of Group 1, with its inputs reduced in magnitude by a factor of $\sqrt{\varepsilon}$. This type of scale heterogeneity can naturally arise when data from different sources record the same underlying attributes using different units or measurement scales. For example, two hospitals may collect the same patient features but record them using different units.

For simplicity of exposition, we also assume there is no label noise and that the shared ground-truth vector $w_*$ has all components equal aligned with the eigendirections $v_1,\ldots,v_d \in \mathbb R^d$ of $\Sigma$, i.e.
\begin{eqnarray}
    \sigma_1 = \sigma_2 = 0,\quad v_j^\top w_* = 1/\sqrt d,\text{ for all }j \in [d].
\end{eqnarray}
For any $t\ > 0$, define the scalar $\eta(t) > 0$ as the unique positive solution to the following fixed-point equation:
\begin{eqnarray}
    \ntrace \Sigma(t \Sigma + \eta I_d)^{-1} = 1.
\end{eqnarray}
The following is one of our main results:
\begin{theorem}
Let the covariance matrices of the groups be as in \eqref{eq:cov-proportional} and suppose $\phi > 1$. Then, in the limit \eqref{eq:proportionate}, followed by $\lambda \to 0^+$ and $\varepsilon \to 0^+$ in that order, the bias error terms in \eqref{eq:bias-variance} are given by $B_1=B$ and $B_2 \simeq \varepsilon B$, where $B=B(p_1,\phi,\psi,\varepsilon)$ is defined by:
\begin{eqnarray}
 B \simeq  \begin{cases}
        \dfrac{p_1\eta(\psi/\phi)}{p_1-\psi},&\mbox{ if }0 < \psi < p_1,\\[9pt]
        \dfrac{\eta(p_1/\phi)}{2}\sqrt{\dfrac{p_1/p_2}{\varepsilon}},&\mbox{ if }\psi=p_1,\\[9pt]
\dfrac{p_2\,\psi \eta(\psi/\phi)}{(\psi - p_1)(1-\psi)},&\mbox{ if }p_1 < \psi < 1\\[9pt]
\infty,&\mbox{ if }\psi = 1,\\[9pt]
\dfrac{\psi \eta(1/\phi)}{\psi-1},&\mbox{ if }\psi > 1.
    \end{cases}  
\end{eqnarray}
\label{thm:triple}
\end{theorem}
Thus, in addition to the classical interpolation singularity at $\psi_+:=1$, the limit $\varepsilon\to0^+$ exhibits an additional singularity at $\psi_-:=p_1$, despite the absence of nonlinear activations. For finite but sufficiently small $\varepsilon$, this additional singularity becomes a rounded peak located at
$\psi=p_1+O(\varepsilon)$, or equivalently,
$
m=n_1+O(\varepsilon n)
$
where $p_1$ is held fixed. Thus, covariance scale separation produces an additional peak near the sample size of the higher-variance group, suggesting an effective interpolation transition associated with fitting that group. More generally, the relevant dimension scales are $\psi_+:=\min(\phi,1)$ and $\psi_-:=\min(\phi,p_1)$. Since $0<p_1<1$, their ordering satisfies
\[
\psi_-\le\psi_+,
\quad
\psi_-<\psi_+
\Longleftrightarrow\quad
p_1<\phi
\Longleftrightarrow\quad
n_1<d.
\]
The scales are therefore distinct precisely when the higher-variance group contains fewer samples than the input dimension. In the regime $\phi>1$ considered by the theorem, this condition holds automatically, giving $\psi_-=p_1<1=\psi_+$. When $\phi\le1$, the upper scale $\psi_+=\phi$ corresponds to feature-rank saturation at $m=d$ and is
not generally an interpolation singularity. The strict ordering identifies when the two scales are separated; the emergence of a rounded peak additionally relies on the spectral and small-$\varepsilon$ conditions of the theorem, producing a triple-descent curve.

\paragraph{Multiple-descent with arbitrarily many peaks.} We now extend our two-group construction, and consider a mixture of $K$ groups with increasingly separated scales. For $k \in [K]$, suppose the covariance matrix of Group $k$ is of the form
\begin{align}
\label{eq:cov-proportional-K}
    \Sigma_k &= \varepsilon_k \Sigma,
\end{align}
where $0 < \varepsilon_K \ll \varepsilon_{K-1} \ll \ldots \ll \varepsilon_k \ll \ldots \ll \varepsilon_1 = 1$. Furthermore, define the cumulative mixture proportion
 $p_{\leq k} := \sum_{s=1}^k p_s
 $
and the corresponding cumulative sample size
\begin{align}
 n_{\leq k} := \sum_{s=1}^k n_s.
\end{align}

\begin{theorem}[K-Fold Multiple Descent]
Given a collection of K groups with covariance structure defined as in \eqref{eq:cov-proportional-K}, for any $k \in [K]$ such that $p_{\leq k} < \phi$, the test error $\mathcal{E}_1$ of the linear random-features model evaluated on Group 1, exhibits $k$ distinct interpolation peaks, occurring at model sizes $m=n_{\leq 1}, n_{\leq 2}, \ldots, n_{\leq k}$. In particular, if $\phi > 1$ (i.e. more input dimensions than total number of training samples),
then $\mathcal{E}_1$ exhibits $K$ distinct interpolation peaks, occurring at model sizes $m=n_{\leq 1},\ldots, n_{\leq K}$.    
\label{thm:K-fold}
\end{theorem}

Multiple descent is thus not limited to a fixed number of interpolation thresholds. Adding a new group (with sufficiently separated covariance scale) to the training mixture introduces a new interpolation peak at the cumulative sample size of the higher-scale groups. Hence, increasing the number of heterogeneous data sources can produce multiple-descent curves with arbitrarily many peaks. 

\paragraph{Triple-descent from heterogeneous power-law covariance spectra.}
We now specialize a setting which characterizes language modeling but also classical learning under so-called capacity constraints, wherein the base covariance matrix $\Sigma$ in \eqref{eq:cov-proportional} has eigenvalues which decay as follows
\begin{eqnarray}
    \lambda_j = c/j,\text{ for all }j \in [d], \text{ and some }c  > 0.
    \label{eq:Zipf}
\end{eqnarray}
\begin{corollary}
Consider the setting of proportional covariance matrices \eqref{eq:cov-proportional} with $\Sigma$ given as in \eqref{eq:Zipf} and suppose $\psi > 1$. Take the limits \eqref{eq:proportionate} followed by $\lambda \to 0^+$. Then, for small $\varepsilon > 0$ the bias terms $B_1$ and $B_2$ for both groups exhibit triple-descent as functions of the parametrization rate $\psi$: each has exactly two local minimal separated by a local maximum at $\psi = p_1 + O(\varepsilon)$ below the classical interpolation threshold at $\psi=1$, increases at the new peak, 
and decreases thereafter. 
Moreover, with noiseless labels, the same conclusions hold for the test error.
\label{corr:Zipf}
\end{corollary}

Corollary~\ref{corr:Zipf} therefore identifies a new type of triple descent since the additional peak emerges without label noise or nonlinear feature maps, but instead from heterogeneity in the covariance scale of the training sources. As $\varepsilon$ becomes small, the two interpolation scales separate, with one peak occurring near the group-specific threshold $m=n_1$ and the classical peak remaining at $m=n$, as illustrated in Figure \ref{fig:triple-cartoon}. A sufficient condition for this behavior is $n_1 > H_{d,1}^2/H_{d,2}$, where $H_{d,\alpha}:=\sum_{j=1}^d j^{-\alpha}$. Since $H_{d,2} = \Theta(1)$ and $H_{d,1} \asymp \log d$, this condition is automatically satisfied in the proportional scaling regime \eqref{eq:proportionate}. 

\paragraph{Implications for Algorithmic Fairness.} Our results also have implications for algorithmic fairness when heterogeneous data sources correspond to different population groups. Prior work has shown that model scaling and inductive biases can amplify group-level performance disparities through mechanisms such as spurious correlations and differences in group difficulty, while heterogeneous training sources can also interact with model size to degrade performance \citep{sagawa2020investigation, bell2023simplicity, arjun2025}. Our results identify covariance heterogeneity as a mechanism for substantial disparities in group-wise prediction error, even with a shared ground-truth predictor and noiseless labels. In the proportional-covariance setting, we have $\mathcal E_2=\varepsilon \mathcal E_1$ so the ratio between the groups' errors becomes arbitrarily large as $\varepsilon\to0^+$. Near the additional peak, this disparity becomes particularly pronounced, with $\mathcal{E}_1$ growing as $\varepsilon^{-1/2}$ while $\mathcal{E}_2$ decreases as $\varepsilon^{1/2}$. This can be empirically observed in Figure \ref{fig:mnist-cifar-scale-exp}. A similarly conclusion carries over to the case where the covariance matrices for the different groups have power-law spectra with different exponents (Figure \ref{fig:mnist-cifar-exponent-exp}). Thus, the same model can perform very differently across groups, with disparities becoming large at certain model sizes, highlighting the importance of group-wise evaluation when training on heterogeneous data.

\section{Experiments}
\label{sec:exp}
We first validate our theoretical predictions on synthetic Gaussian data mixtures, then test whether the same qualitative phenomena persist on real-world datasets such as MNIST \citep{deng2012mnist} and CIFAR-10 \citep{krizhevsky2009cifar}. 

\subsection{Synthetic Data}
\label{sec:synthgaussdata}

\paragraph{Setup.} We consider mixtures of Gaussian data generated according to \eqref{eq:gaussinputs}--\eqref{eq:gausslabels}, allowing us to isolate the effects of covariance heterogeneity: differences in spectral structure and differences in overall covariance scale. Throughout the experiments, the labels are noiseless (i.e. $z_1 = z_2 = 0$) and all groups share the same ground-truth weights $w_* \in \mathbb{R}^d$ that is sampled uniformly from the sphere of radius $\sqrt{d}$. 

\paragraph{Heterogeneity in spectral structure.} We first ask if multiple descent can arise when the groups differ in the shape of their covariance spectra. For this experiment, we consider two groups with power-law covariance spectra, defined respectively as $\Sigma_1 = \operatorname{diag}(\lambda_{1}^{(1)},...,\lambda_{d}^{(1)})$ and $\Sigma_2 = \operatorname{diag}(\lambda_1^{(2)},...,\lambda_d^{(2)})$, where the $\lambda_j$'s satisfy:
\begin{eqnarray}
\lambda_j^{(1)} \propto j^{-\alpha},\quad \lambda_j^{(2)} \propto j^{-\beta},\text{ for all }j \in [d],\text{ and constants }0 < \alpha \leq 1,\quad \beta > \alpha + 1.    
\end{eqnarray}
The conditions on the exponents make Group 1 have substantially flatter covariance spectrum than Group 2. In our experiment, we use $\alpha = 1$ and $\beta = 2.2$. We set $d=2000$ and fix the total number of training samples at $n=1000$, varying the proportion $p_1 = n_1 / n \in \{1/4, 1/2, 3/4\}$ so that $n_1 \in \{250, 500, 750\}$ and $n_2 = n - n_1$. We fit a linear random-features model $\hat{f}$ according to \eqref{eq:linearrf} (with linear activation function) for different model sizes $m$, and set the regularization to a small value  $\lambda = 10^{-12}$ (which is essentially zero). 

\begin{figure*}[!htp]
\centering
\includegraphics[width=.48\linewidth]{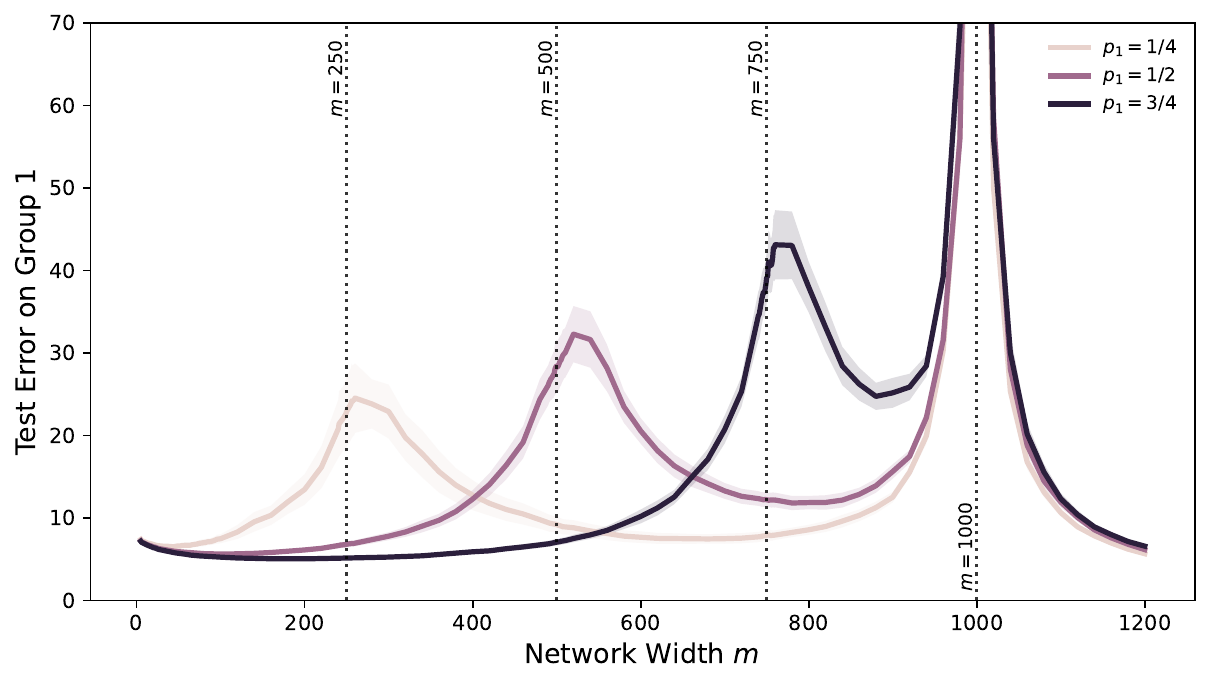}
\includegraphics[width=.48\linewidth]{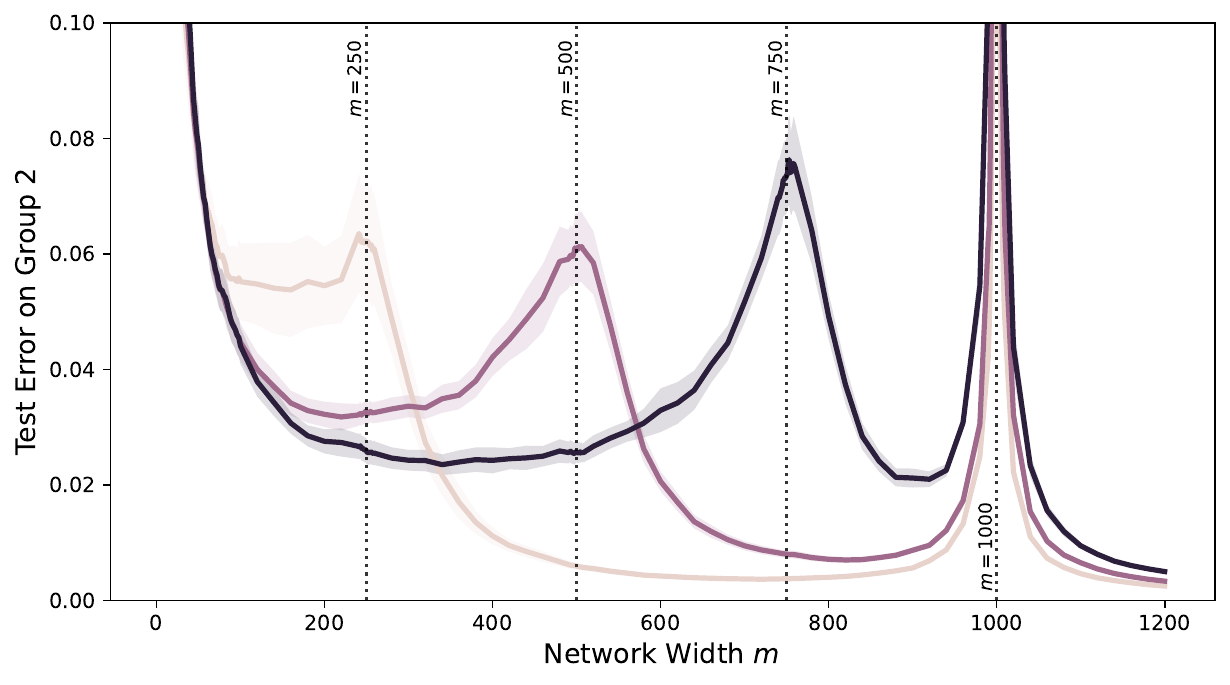}
\vspace{-.3cm}
\caption{\textbf{Triple-descent in linear RF on  synthetic data mixture with different spectral exponents.} We train a linear  RF model \eqref{eq:linearrf} on a mixture of two Gaussian groups \eqref{eq:gaussinputs}--\eqref{eq:gausslabels} with total sample size $n=1000$. Group 1 has covariance matrix $\Sigma_1=\operatorname{diag} (\lambda_1^{(1)},\ldots,\lambda_d^{(1)})$, where $\lambda_i^{(1)}\propto i^{-\alpha}$ and $\alpha=1$. Group 2 is defined analogously with $\lambda_i^{(2)}\propto i^{-\beta}$ and $\beta=2.2$. Thus, Group 1 has the flatter spectrum and corresponds to the harder-to-learn population. We plot the test error of Group 1 (\textbf{left}) and Group 2 (\textbf{right}) versus the model's width $m$ for different mixture proportions $p_1 \in \{1/4, 1/2, 3/4\}$, over 20 runs (error bars). In addition to the classical interpolation threshold at $m=1000$, each curve exhibits an additional peak near the group-specific threshold $m=n_1=p_1 n$ as predicted by Corollary \ref{corr:Zipf}.}
\label{fig:triple-gaussian-exp}
\end{figure*}

Figure \ref{fig:triple-gaussian-exp} shows the test error on Group 1 (\textbf{left}) and Group 2 (\textbf{right}) as a function of model size $m$ for the three mixture proportions $p_1 \in \{1/4, 1/2, 3/4\}$. As the model width increases, both groups exhibit a first interpolation peak near the group-specific threshold $m=n_1=p_1 n$, followed by a second peak at the classical interpolation threshold $m=n=1000$. As $p_1$ increases, the first peak shifts from $m=250$ to $m=500$ to $m=750$, while the second peak remains fixed at $m=1000$. Thus, changing the composition of the training mixture changes the location of the additional interpolation threshold without affecting the classical one. Beyond $m=n$, the test error decreases again for both groups. Hence these results agree with the qualitative prediction of our theory that heterogeneity in the spectral exponents of the two data sources 
is sufficient to produce the classical interpolation threshold at $m=n$ and the additional peak at $m=n_1=p_1 n$.

\paragraph{Heterogeneity in covariance scale.} We next ask whether multiple descent can arise if the groups differ in their overall scale rather than only in their shape. To isolate scale heterogeneity from differences in covariance shape, we consider $K=4$ isotropic Gaussian groups with covariance matrices defined as $\Sigma_k = \varepsilon_k I_d$ ($\text{for } k \in [K]$), where we set the scales as $(\varepsilon_1, \varepsilon_2, \varepsilon_3, \varepsilon_4) = (1, 10^{-3}, 10^{-5}, 10^{-7})$. We use group sizes $(n_1, n_2, n_3, n_4) = (50, 100, 200, 250)$, giving a total sample size of $n=600$. As before, all groups share the same noiseless linear teacher. 

\begin{figure}[!t]
\centering
\includegraphics[width=0.70\linewidth]{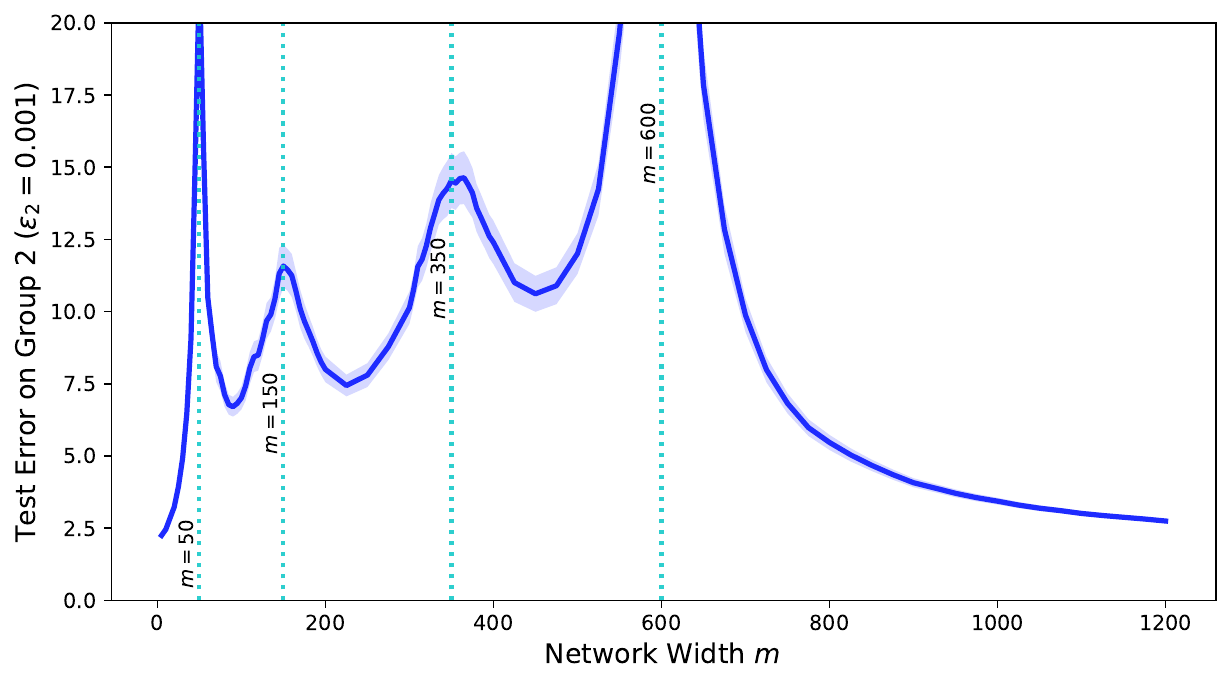}
\vspace{-.3cm}
\caption{\textbf{Multiple-descent in linear RF on synthetic data mixture with different scales.} We train a linear RF model \eqref{eq:linearrf} on a mixture of four Gaussian groups with total sample size $n = 600$. For Group k, the inputs are sampled as $x^{(k)} \sim \mathcal{N}(0, \Sigma_k)$, with isotropic covariance $\Sigma_k = \varepsilon_k I_d$. The four groups have sizes $(n_1, n_2, n_3, n_4) = (50, 100, 200, 250)$ and scales $(\varepsilon_1, \varepsilon_2, \varepsilon_3, \varepsilon_4) = (1, 10^{-3}, 10^{-5}, 10^{-7})$. We plot the test error of Group 2 versus the model's width $m$ over 20 runs (error bars). As predicted by Theorem \ref{thm:K-fold}, the test error of Group 2 exhibits multiple interpolation peaks near the cumulative sample sizes $m=n_1=50$, $m=n_1+n_2=150$, $m=n_1+n_2+n_3=350$, and $m=n=600$. An extended figure with test error curves for all 4 groups is presented in Appendix \ref{sec:additional-experiments-1}.}
\label{fig:group2-multiple-gaussian-exp}
\end{figure}

\begin{figure*}[!htp]
\centering
\includegraphics[width=.48\linewidth]{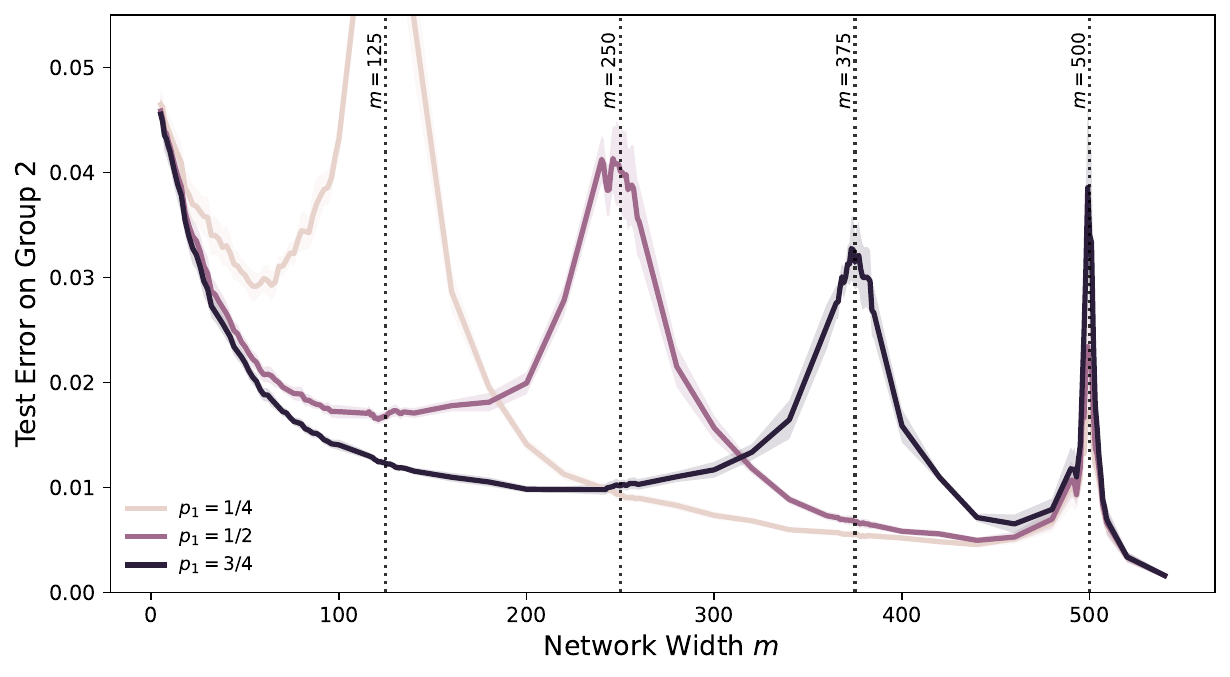}
\includegraphics[width=.48\linewidth]{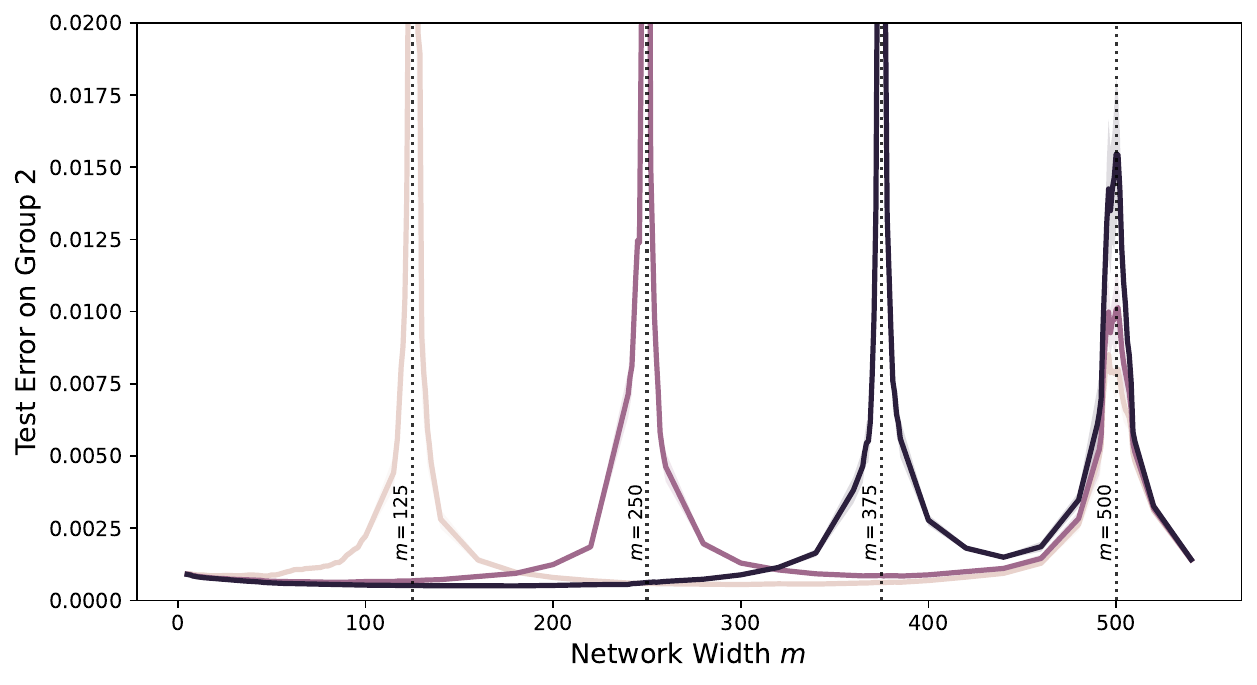}
\vspace{-.3cm}
\caption{\textbf{Triple-descent in linear RF model on MNIST \& CIFAR-10 mixtures with different scales.} We train linear RF models \eqref{eq:linearrf} on two-group mixtures constructed from MNIST and CIFAR-10. For each dataset, Group 1 consists of the original inputs, while Group 2 is obtained by re-scaling the inputs by $\sqrt{\varepsilon}$, so that $\Sigma_2 = \varepsilon \Sigma_1$. We use $\varepsilon = 10^{-3}$ for MNIST and $\varepsilon = 10^{-5}$ for CIFAR-10. In all experiments, the total training sample size is fixed at $n=500$, and we vary the mixture proportion $p_1 \in \{1/4, 1/2, 3/4\}$. We plot the test error of Group 2 derived from MNIST (\textbf{left}) and derived from CIFAR-10 (\textbf{right}) as a function of model size $m$, over 10 runs (error bars). For both groups, an additional interpolation peak appears near the group-specific threshold $m=n_1=p_1 n$, shifting with the mixture proportion, while the classic interpolation peak remains fixed at $m=n=500$. An extended figure (with Group 1 for both datasets) can be found in Appendix \ref{sec:additional-experiments-2}.}
\label{fig:mnist-cifar-scale-exp}
\end{figure*}

\begin{figure*}[!htp]
\centering
\includegraphics[width=.48\linewidth]{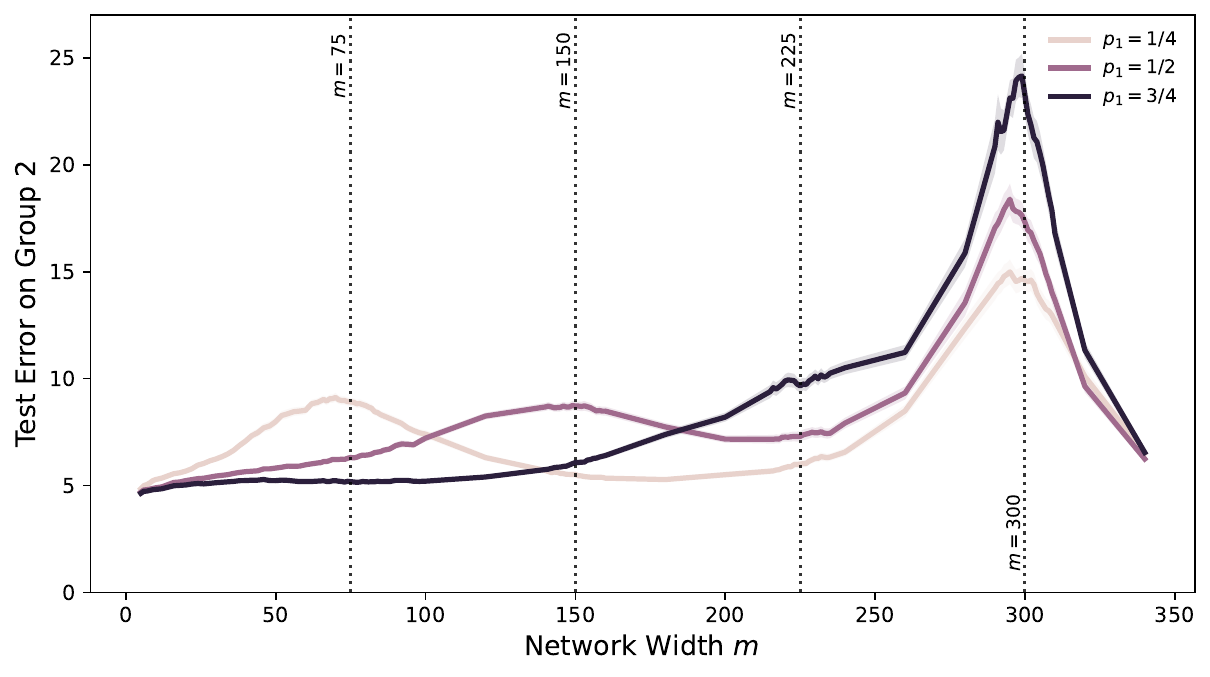}
\includegraphics[width=.48\linewidth]{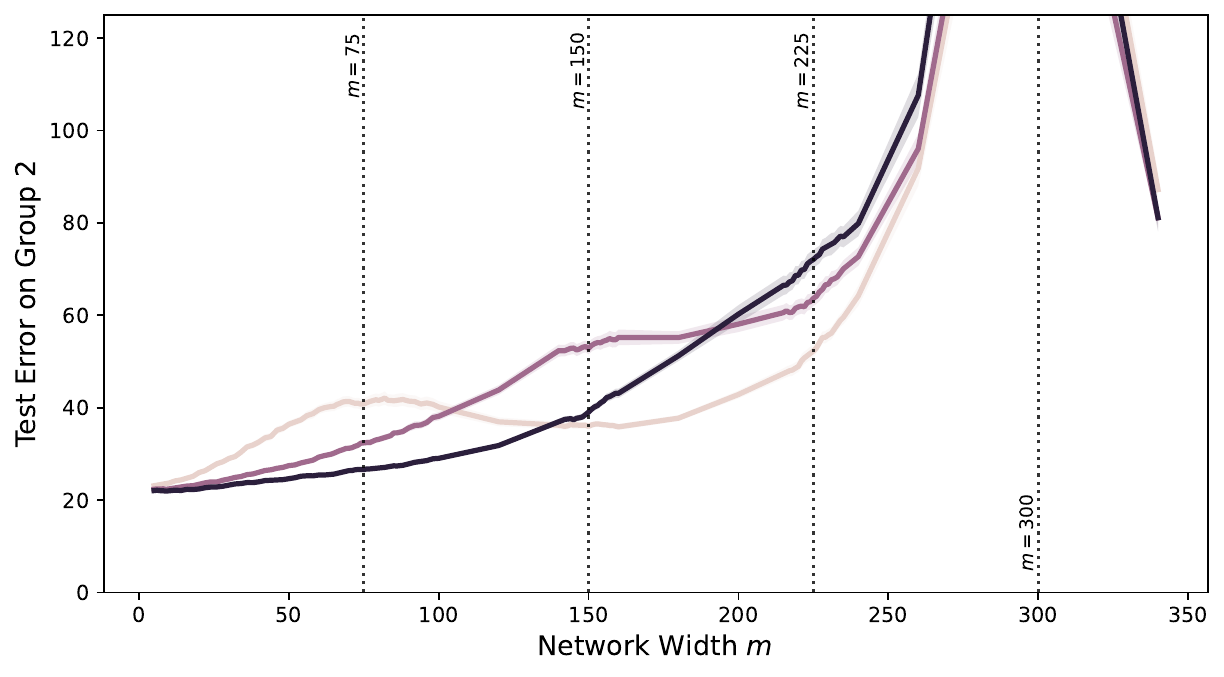}
\vspace{-.3cm}
\caption{\textbf{Triple-descent in linear RF model on MNIST \& CIFAR-10 mixtures with different exponents.} We train linear RF models \eqref{eq:linearrf} on two-group mixtures constructed from MNIST and CIFAR-10. For each dataset, Group 1 consists of the original inputs, while Group 2 is obtained through a nonlinear transformation that produces a flatter empirical covariance spectrum. All experiments fix $n=300$ and vary $p_1 \in \{1/4,1/2,3/4\}$. We plot the test error of Group 2 derived from MNIST (\textbf{left}) and derived from CIFAR-10 (\textbf{right}) as a function of model size $m$, over 50 runs for MNIST and 30 runs for CIFAR-10 (error bars). In both groups, an additional peak appears near the group-specific threshold $m=n_1=p_1 n$ predicted by our theory (Corollary~\ref{corr:Zipf}), while the classical interpolation peak remains fixed at $m=n=300$. Additional results are provided in Appendix \ref{sec:additional-experiments-2}.  
}
\label{fig:mnist-cifar-exponent-exp}
\end{figure*}

Figure \ref{fig:group2-multiple-gaussian-exp} shows the test error of Group 2 as the model size $m$ increases. In contrast to our previous experiment, the curves now exhibit a sequence of interpolation peaks at $m=n_1=50$, $m=n_1+n_2=150$, $m=n_1+n_2+n_3=350$ and $m=n=600$, which agrees with our theoretical predictions (Theorem~\ref{thm:K-fold}). This demonstrates that strong heterogeneity in covariance scale can generate arbitrarily many interpolation peaks.
See Appendix~\ref{sec:additional-experiments-1} for plots for the other groups.

\subsection{Real-World Data}
\label{sec:real-world-data}

\paragraph{Setup.} We next investigate whether the results obtained from Section~\ref{sec:synthgaussdata} extend to settings where the Gaussian input assumption is replaced by inputs from the MNIST and CIFAR-10 datasets. As before, we consider two groups with total training size $n = n_1 + n_2$, and vary the mixture proportion $p_1 = n_1 / n$. For the labels, we omit label noise (i.e. $z_1 = z_2 = 0$) and again all groups share the same ground-truth weights $w_* \in \mathbb{R}^d$ that is sampled uniformly from sphere of radius $\sqrt{d}$. We fit a linear random-features model $\hat{f}$ according to \eqref{eq:linearrf} across a range of model widths $m$. 

\paragraph{Impact of covariance scale.} We first study the proportional-covariance setting of \eqref{eq:cov-proportional}. For each dataset, Group 1 consists of the original inputs and Group 2 is obtained by scaling a separate set of samples by $\sqrt{\varepsilon}$, so that $\Sigma_2 = \varepsilon \Sigma_1$. For both groups, we fix the total training size to $n=500$ and vary $p_1 \in \{1/4, 1/2, 3/4\}$. We use $\varepsilon=10^{-3}$ for MNIST and $\varepsilon=10^{-5}$ for CIFAR-10, and set the regularization parameter to $\lambda=10^{-12}$ in both cases. 

Figure \ref{fig:mnist-cifar-scale-exp} shows that for both groups, an additional interpolation peak appears near $m=n_1=p_1 n$ and shifts with the mixture proportion, while the classical interpolation peak remains fixed at $m=n=500$. This provides evidence that the scale induced threshold identified by our theory persists beyond the Gaussian setting, to real-world heterogeneous datasets. Group 1 results are reported in Appendix \ref{sec:additional-experiments-2}. 

\paragraph{Impact of spectral structure.} We next consider groups with different covariance spectral shapes. For each dataset, Group 1 consists of the original inputs, while Group 2 is obtained by applying a nonlinear transformation $T(x)$ to the original inputs, and chosen so that it produces a flatter empirical covariance spectrum. We estimate the spectrum of each group and fit a power law of the form $\lambda_i \propto i^{-s}$, where $s$ denotes the spectral exponent. The trace-normalized spectra and corresponding power-law fits are shown in Appendix~\ref{sec:additional-experiments-2}. For both MNIST and CIFAR-10, the two groups exhibit distinct rates of spectral decay.

For both MNIST and CIFAR-10, we fix $n=300$, vary $p_1 \in \{1/4,1/2,3/4\}$, corresponding to $n_1 \in \{75,150,225\}$, and set the regularization parameter to $\lambda = 10^{-6}$. Figure~\ref{fig:mnist-cifar-exponent-exp} shows the resulting Group 2 test errors. As the mixture proportion varies, an additional peak appears at $m=n_1=p_1 n$, while the classical interpolation peak remains fixed at $m=n=300$. The same qualitative behavior is observed for both MNIST and CIFAR-10, showing that multiple interpolation peaks can arise when the groups differ in their empirical spectral structure. The corresponding Group 1 errors are provided in Appendix~\ref{sec:additional-experiments-2}.

\section{Related Work}
\label{sec:related-work}
\textbf{Double-descent.} The phenomenon of (model-wise) double descent, where a model's test error decreases, then increases near the interpolation threshold, and decreases again as the model becomes increasingly over-parameterized, was first introduced to the machine learning community by \citet{belkin2019reconciling}, although similar behavior had been observed in the works of \citet{opper1995statistical} and \citet{advani2020high}. A substantial theoretical literature has since studied model-wise double descent in a tractable linear least-squares settings \citep{belkin2019reconciling, bartlett2020benign, mei2022generalization, Hastie2019Surprises, bach2024high}. The works of \citet{nakkiran2021deep} and \citet{pezeshki2022multi} study the less explored variant of epoch-wise double descent, where the test error decreases, then increases, and decreases again as training time increases. The former demonstrates this behavior empirically across several neural network settings and interprets it through the notion of effective model complexity, while the latter provides a theoretical explanation in a linear teacher-student model, showing that epoch-wise double descent can arise when different features are learned at different time scales.

\textbf{Triple- and multiple-descent.} More complex generalization curves exhibiting multiple interpolation peaks have also been observed \citep{d2020triple, adlam2020neural}. The work of \citet{d2020triple} studies sample-wise triple descent in random-feature models theoretically and in deep neural networks empirically, showing that two distinct peaks can appear due to label noise and nonlinearity of the activation function. The work of \citet{adlam2020neural} extends previous theoretical analyses beyond linear regression and kernel regression with unstructured random features by studying high-dimensional kernel regression with the Neural Tangent Kernel (NTK). They show that the test error can exhibit additional non-monotonic behavior deep in the overparameterized regime, with a second peak emerging when the number of model parameters scales quadratically with the number of training samples, in addition to the usual interpolation peak associated with linear scaling. In contrast, our work shows that multiple descent behavior can arise in linear random-feature models with noiseless labels solely from heterogeneity in the covariance structures of the training data sources.

\textbf{Learning from heterogeneous data.} Recent theoretical work has studied how heterogeneity across training data sources can affect generalization \citep{dohmatob2024tale, dohmatob2024modelcollapse, bertrand2024stability,gerstgrasser2024is,dohmatob2025strong, barzilai2026models} The work of \citet{dohmatob2025strong} analyzes mixtures of real and synthetic data in random-projections models, showing that even a small fraction of synthetic data can lead to persistent model collapse \citep{shumailov2024aimodelscollapse}, and that the severity of this effect depends strongly on model size. \citet{arjun2025} develop a high-dimensional theory of bias amplification \citep{zhao2017men, hendricks2018women, wang2021directional} for models trained on mixtures of groups with potentially different covariance structures, label noise levels, and ground-truth predictors, characterizing how these differences interact with model size to produce disparities in group-wise test error. Although their framework permits group-dependent covariance structures, their focus is on bias amplification across groups rather than on the interpolation structure induced by covariance heterogeneity. Our work instead studies how differences in the statistical structure of the data sources determine the number and location of interpolation thresholds as model size varies. Nevertheless, the phenomena studied in these earlier works correspond to particular regimes of our general heterogeneous-data framework, allowing several of their results to be recovered as special cases.

\section{Concluding Remarks}
We establish covariance heterogeneity as a mechanism for multiple descent that requires neither label noise nor nonlinear activations. Our asymptotically exact source-specific test error formulas reveal how covariance scale and spectral structure generate additional error peaks and pronounced group-wise disparities as model size varies. Experiments on synthetic data, MNIST, and CIFAR-10 confirm these predictions and demonstrate their relevance beyond Gaussian inputs.

Our findings broaden the classical picture of scaling models up: the statistical composition of training data can itself create distinct learning scales, making source-specific evaluation important for understanding how model scaling affects different groups.

\paragraph{Limitations and Future Directions.}
Our theoretical guarantees assume Gaussian inputs and proportional high-dimensional scaling. Future work could extend these guarantees to broader input distributions. A complementary direction is to develop principled training strategies that reduce group-wise error disparities as models scale, connecting the theory of generalization with algorithmic fairness.

\subsection*{Acknowledgement} We would like to proudly acknowledge that the first author \textbf{Mark-Anthony Moisescu-Pareja} is an MSc student funded by bursaries from Canada Research Chair grant number CRC-2023-00228 (\emph{Computing for the Greater Good}) awarded to \textbf{Elvis Dohmatob}.

\subsection*{Note on AI Use}
In this work, we have made limited use of generative AI tools for spell-checking, and also reformulating chunks of text in the introduction (Section \ref{sec:intro}), detailed related works section (Section \ref{sec:related-work}), and verifying some proofs (Appendix \ref{sec:var-analytic}). Also, AI was used to generate Tikz code for displaying the cartoon in Figure \eqref{fig:triple-cartoon}. No other use of AI was made in our work. We take full responsibility for the final content of this work, including text, claims, or artifacts produced with the aid of AI.

\bibliography{references}

\appendix
\etocdepthtag.toc{appendix}

\clearpage
\thispagestyle{empty}
\vspace*{\fill}

\begin{center}
    {\Huge\bfseries Appendix}
\end{center}


\begingroup
\etocsettocstyle{\section*{}}{}
\etocsettagdepth{main}{none}
\etocsettagdepth{appendix}{subsection}
\tableofcontents
\endgroup

\newpage

\section{Additional Experiments Results}
\label{sec:additional-experiments}


\subsection{Synthetic Gaussian Data}
\label{sec:additional-experiments-1}

\begin{figure*}[!htp]
\centering
\includegraphics[width=.48\linewidth]{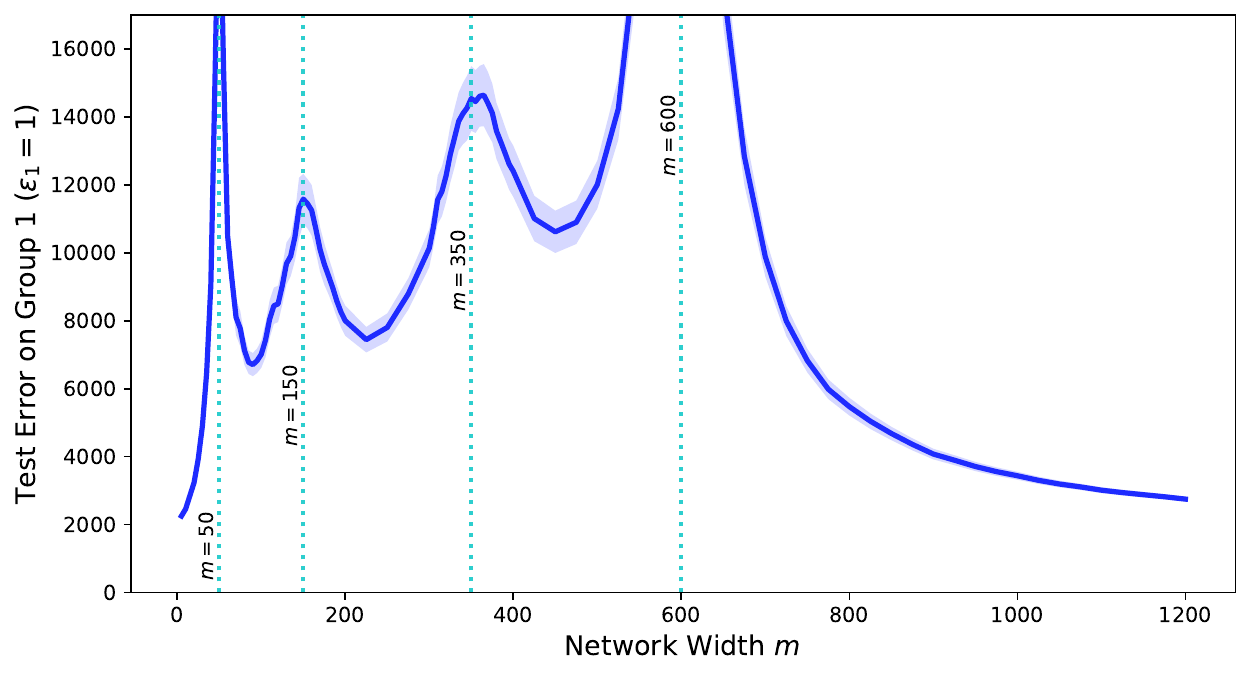}
\includegraphics[width=.48\linewidth]{synthetic_K=4_Group2.pdf}
\includegraphics[width=.48\linewidth]{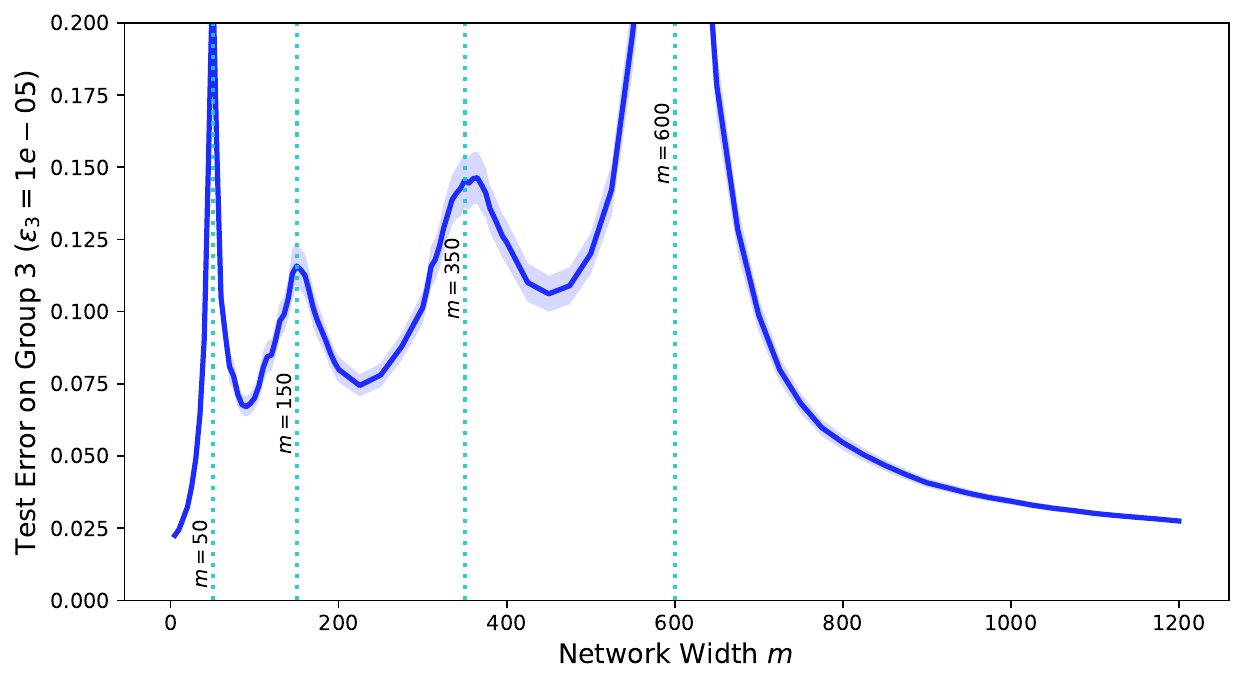}
\includegraphics[width=.48\linewidth]{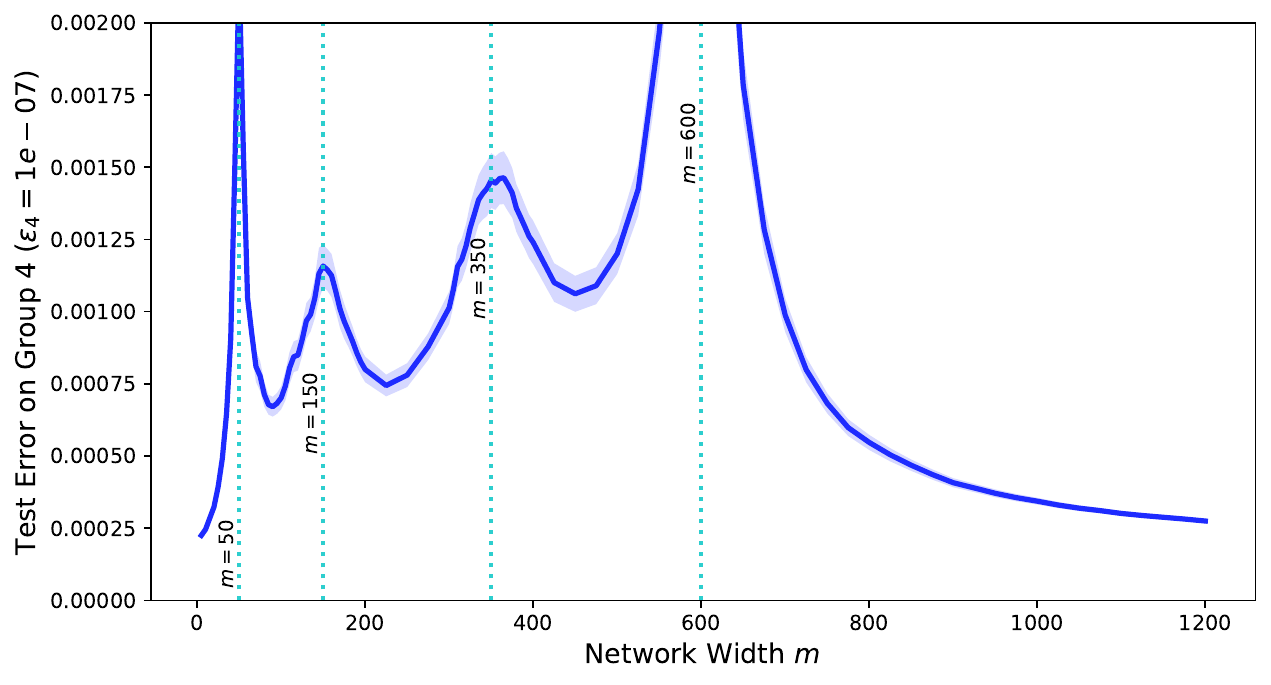}
\vspace{-.3cm}
\caption{\textbf{Multiple-descent in linear RF on synthetic data mixture with different scales.} For this experiment, we train a linear random-features model on a mixture of four Gaussian groups with total sample size $n = 600$. For Group k, the inputs are sampled as $x^{(k)} \sim \mathcal{N}(0, \Sigma_k)$, with isotropic covariance $\Sigma_k = \varepsilon_k I_d$. The four groups have sizes $(n_1, n_2, n_3, n_4) = (50, 100, 200, 250)$ and scales $(\varepsilon_1, \varepsilon_2, \varepsilon_3, \varepsilon_4) = (1, 10^{-3}, 10^{-5}, 10^{-7})$, so that the groups differ only in the overall scale of their input covariances. The labels in each group are generated noiselessly by the same ground-truth weights (i.e. $w_*^{(1)} = w_*^{(2)} = w_*^{(3)} = w_*^{(4)} = w_*$). We plot the test error of Group 1 (\textbf{upper left}), Group 2 (\textbf{upper right}), Group 3 (\textbf{bottom left}), and Group 4 (\textbf{bottom right}) versus the model's width $m$ over 20 runs (error bars). As predicted by Theorem \ref{thm:K-fold}, the test error of each group exhibits multiple interpolation peaks near the cumulative sample sizes $m=n_1=50$, $m=n_1+n_2=150$, $m=n_1+n_2+n_3=350$, and $m=n=600$, demonstrating that strong heterogeneity in covariance scale can generate arbitrarily many interpolation peaks (that is dependent on the number of mixtures in the data source).
}
\label{fig:multiple-gaussian-exp}
\end{figure*}

\newpage

\subsection{Real-World Data}
\label{sec:additional-experiments-2}

\begin{figure*}[!htp]
\centering
\includegraphics[width=1\linewidth]{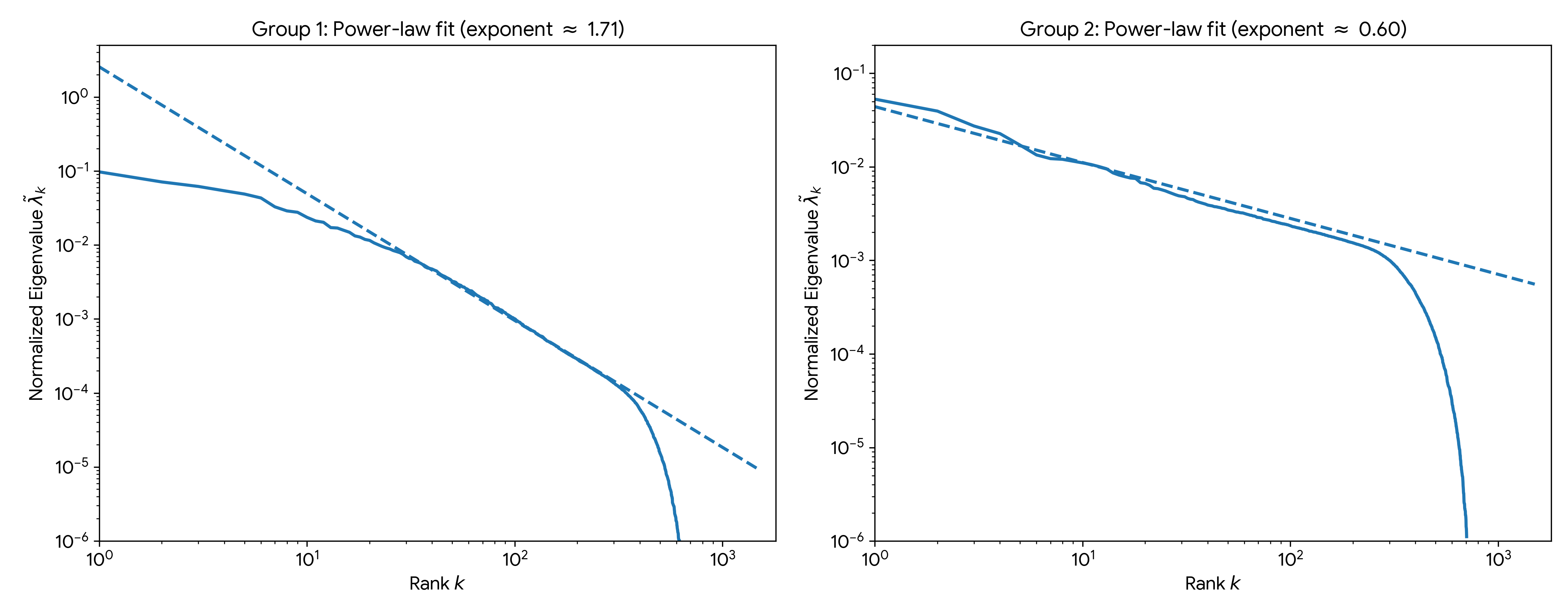}
\vspace{-.5cm}
\caption{\textbf{Power-law spectra for MNIST.} We plot the trace-normalized eigenvalues of the empirical covariance matrices against eigenvalue rank $k$ for Group 1 (original MNIST inputs) and Group 2 (nonlinearly transformed inputs). Solid lines denote the empirical spectra, while dashed lines denote the corresponding power-law fits.}
\label{fig:mnist-spectra-appendix}
\end{figure*}

\begin{figure*}[!htp]
\centering
\includegraphics[width=1\linewidth]{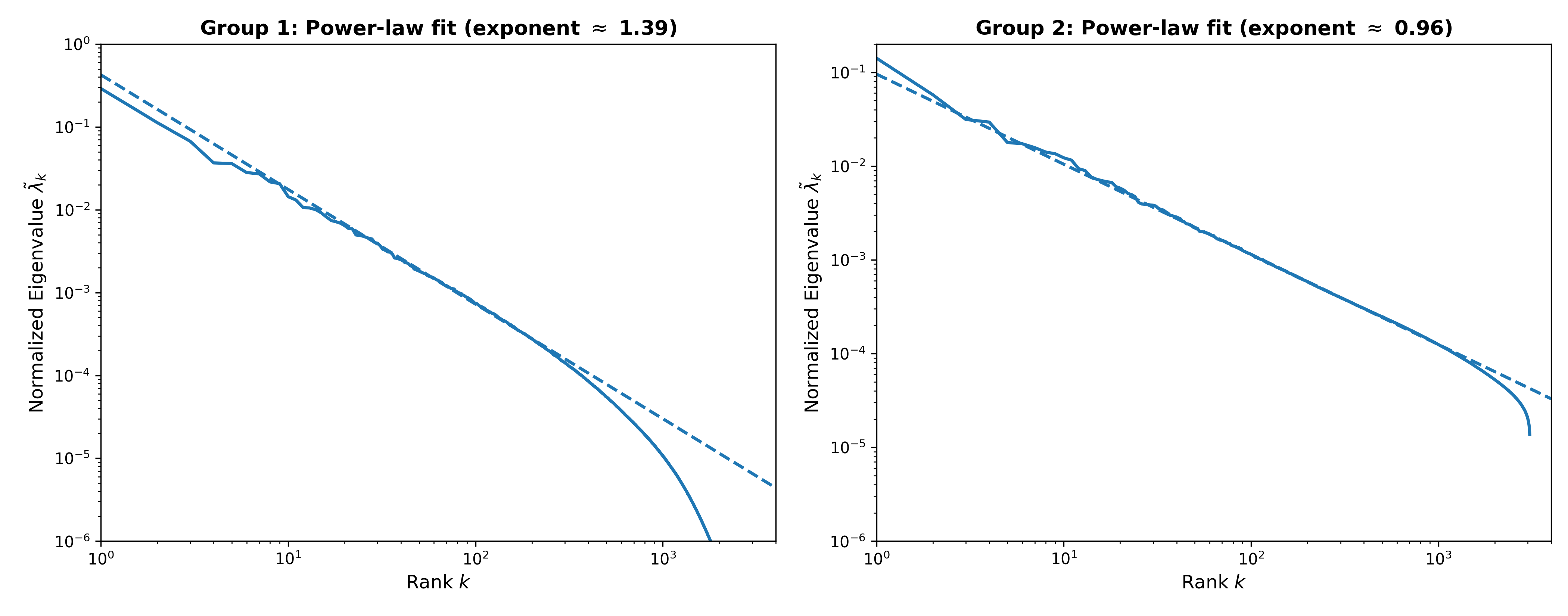}
\vspace{-.5cm}
\caption{\textbf{Power-law spectra for CIFAR-10.} We plot the trace-normalized eigenvalues of the empirical covariance matrices against eigenvalue rank $k$ for Group 1 (original CIFAR-10 inputs) and Group 2 (nonlinearly transformed inputs). Solid lines denote the empirical spectra, while dashed lines denote the corresponding power-law fits.}
\label{fig:mnist-spectra-appendix}
\end{figure*}

\begin{figure*}[!htp]
\centering
\includegraphics[width=.48\linewidth]{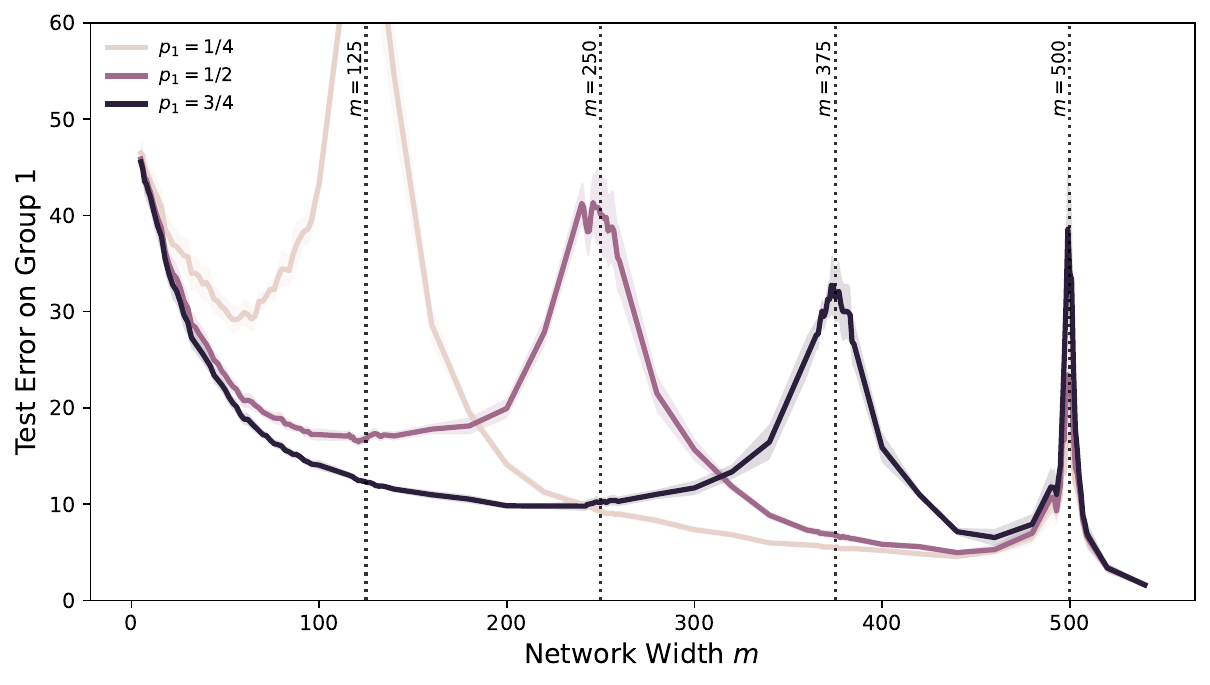}
\includegraphics[width=.48\linewidth]{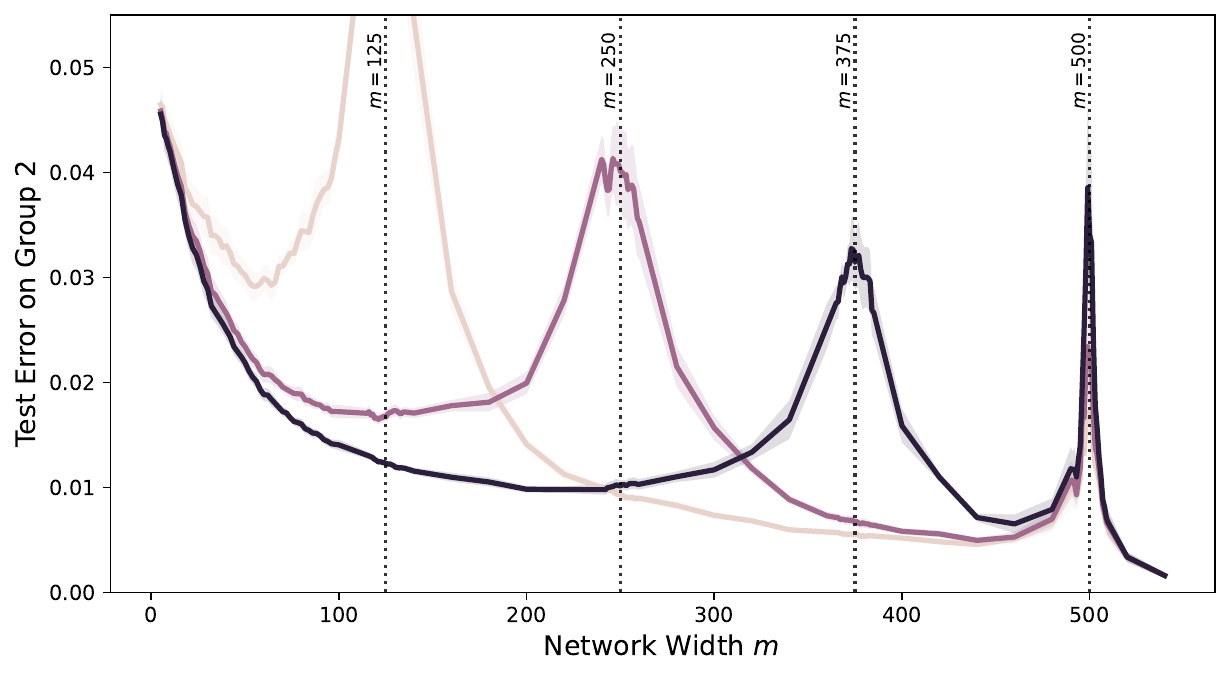}
\includegraphics[width=.48\linewidth]{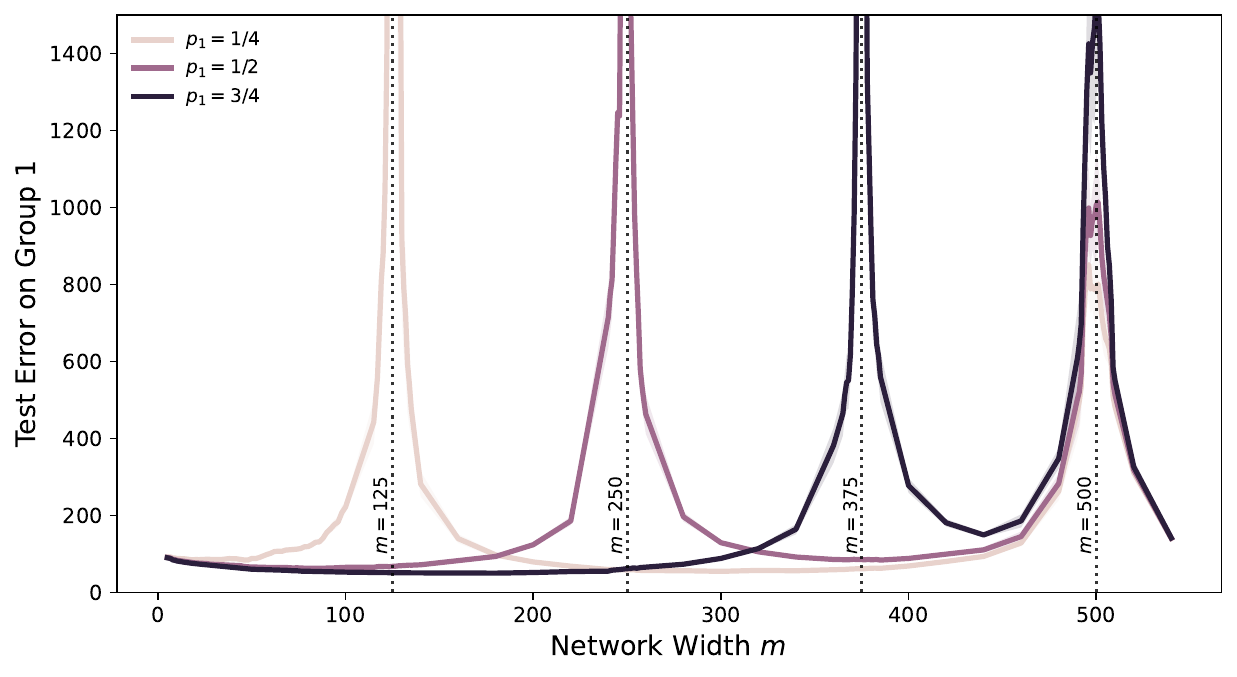}
\includegraphics[width=.48\linewidth]{Group2_CIFAR10_scale_experiment.pdf}
\vspace{-.3cm}
\caption{\textbf{Triple-descent in linear RF on MNIST \& CIFAR-10 mixtures with different scales.} We train linear random-features models on two-group mixtures constructed from MNIST (\textbf{top row}) and CIFAR-10 (\textbf{bottom row}). For each dataset, Group 1 consists of the original inputs, while Group 2 is obtained by re-scaling the inputs by $\sqrt{\epsilon}$, so that $\Sigma_2 = \epsilon \Sigma_1$. We use $\epsilon = 10^{-3}$ for MNIST and $\epsilon = 10^{-5}$ for CIFAR-10. In both experiments, the total training sample size is fixed at $n=500$, and we vary the mixture proportion $p_1 \in \{1/4, 1/2, 3/4\}$, corresponding to $n_1 \in \{125, 250, 375\}$. We plot the test error of Group 1 (\textbf{left column}) and Group 2 (\textbf{right column}) as a function of model size $m$, over 10 runs (error bars). For both MNIST and CIFAR-10, an additional interpolation peak appears near the group-specific threshold $m=n_1=p_1 n$, shifting with the mixture proportion, while the classic interpolation peak remains fixed at $m=n=500$. }
\label{fig:mnist-cifar-scale-exp-appendix}
\end{figure*}

\begin{figure*}[!htp]
\centering
\includegraphics[width=.48\linewidth]{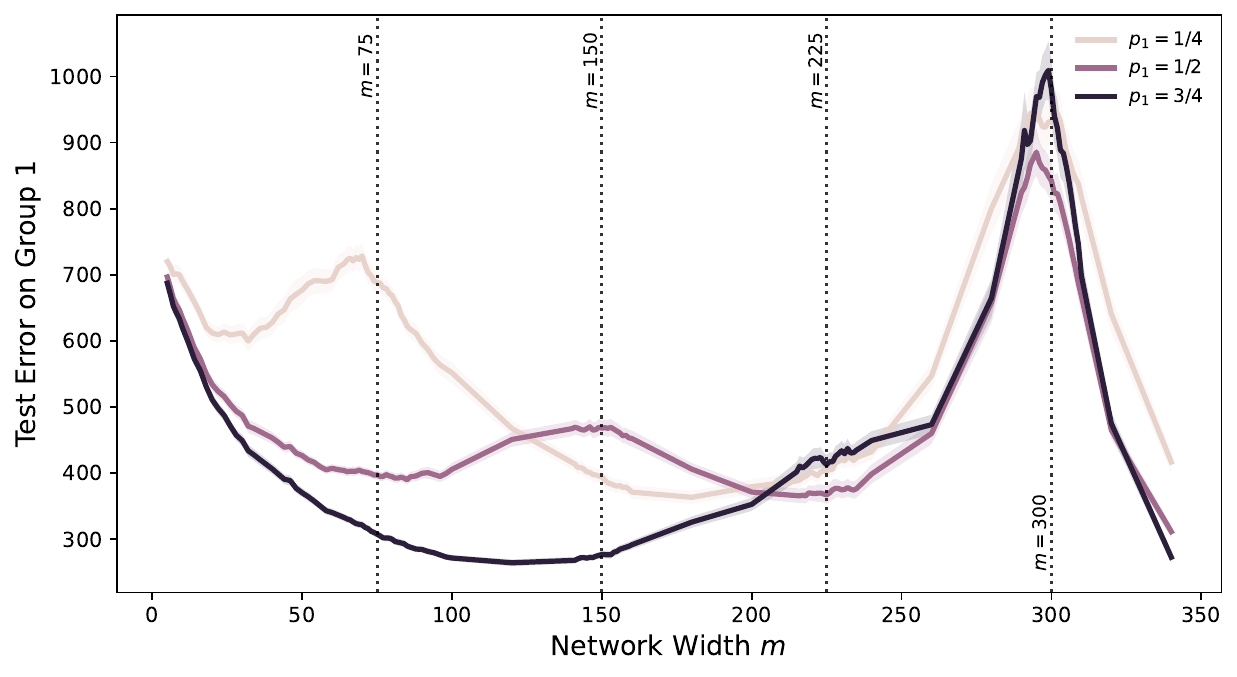}
\includegraphics[width=.48\linewidth]{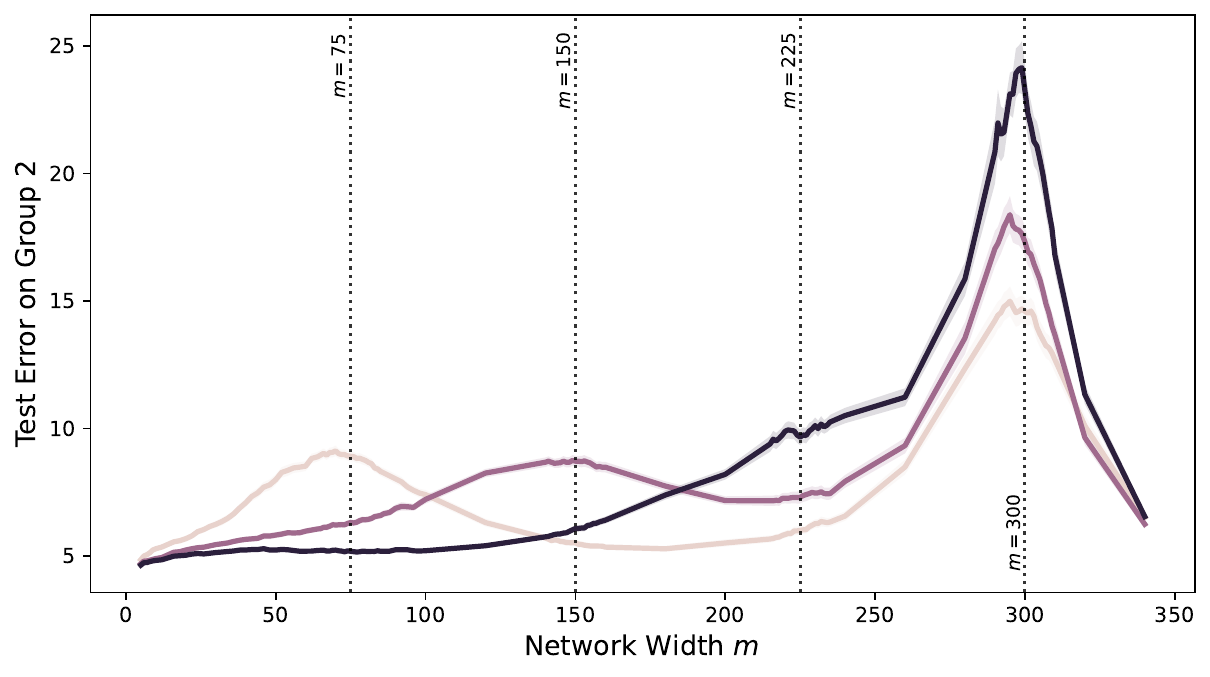}
\includegraphics[width=.48\linewidth]{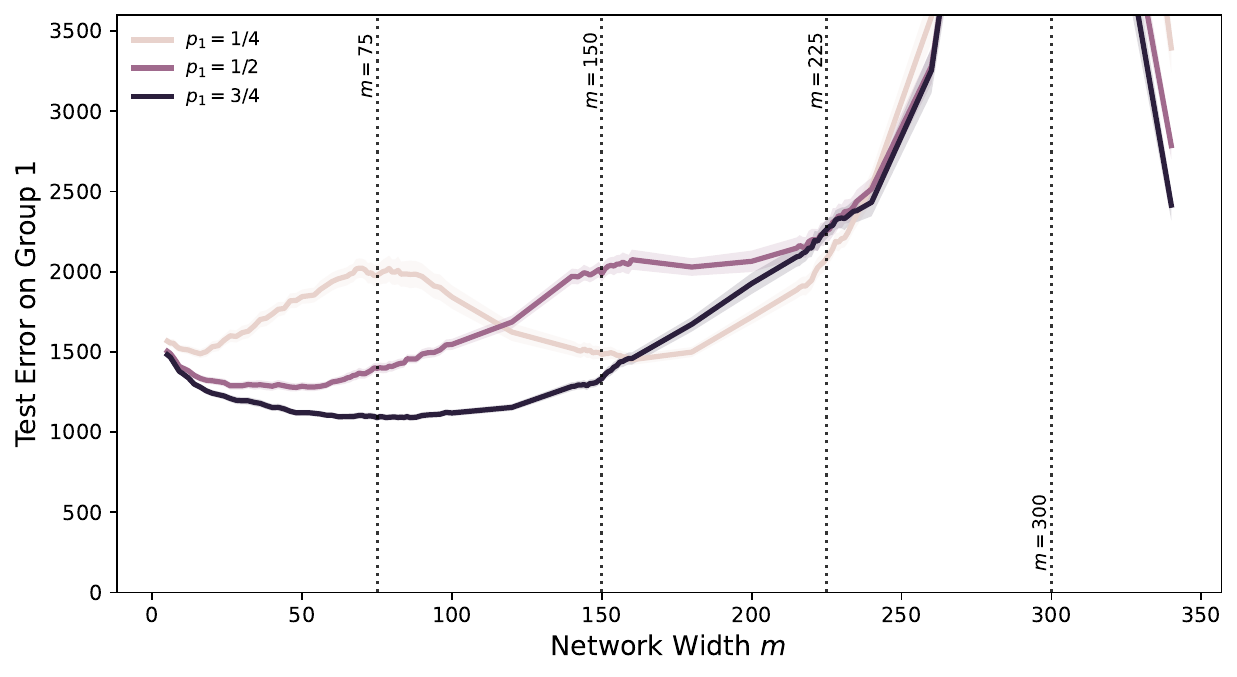}
\includegraphics[width=.48\linewidth]{Group2_CIFAR_exponent_experiment_nolegend_v2.pdf}
\vspace{-.3cm}
\caption{\textbf{Triple-descent in linear RF on MNIST \& CIFAR-10 mixtures with different exponents.} We train linear random-features models on two-group mixtures derived from MNIST (\textbf{top row}) and CIFAR-10 (\textbf{bottom row}). For each dataset, Group 1 consists of the original inputs, while Group 2 is obtained through a nonlinear transformation that produces a flatter empirical covariance spectrum. In both experiments, the total training sample size is fixed at $n=300$, and we vary the mixture proportion $p_1 \in \{1/4,1/2,3/4\}$, corresponding to $n_1 \in \{75,150,225\}$. We plot the test error of Group 1 (\textbf{left column}) and Group 2 (\textbf{right column}) as a function of model size $m$, over 50 runs for MNIST and 30 runs for CIFAR-10 (error bars). For both MNIST and CIFAR-10, an additional interpolation peak appears near the group-specific threshold $m=n_1=p_1 n$ predicted by our theory (Corollary \ref{corr:Zipf}), while the classic interpolation peak remains fixed at $m=300$.}
\label{fig:mnist-cifar-scale-exp-appendix}
\end{figure*}

\clearpage

\section{Bias-Variance Decomposition (Proof of Theorem \ref{thm:main})}
\label{sec:bvproofs}

\subsection{Important Constants}
\label{sec:constants}
Define the following auxiliary quantities:
\begin{eqnarray}
\boxed{
    \begin{split}
   E_i &:= \phi \ntrace \Sigma_i K^{-2}, \quad  \tilde E_i:= \phi \ntrace B \Sigma_i K^{-2}, \quad F_i := \phi \ntrace \Theta \Sigma_i K^{-2},\quad \tilde F_i := \phi \ntrace\Theta B \Sigma_i K^{-2}\text{ for }i \in \{0,1,2\},\\
    D_{ij} &:= \phi\ntrace\Sigma_i \Sigma_j K^{-2},\quad a_i := \rho_i^2/\psi^2 - p_i D_{ii},\quad \check D_{ij} := \phi\ntrace\Theta \Sigma_i \Sigma_j K^{-2}\text{ for }i,j \in \{1,2\},\\
    A &:= \phi\ntrace K_0^2 K^{-2} = \frac{p_1^2 D_{11}}{\rho_1^2} + \frac{p_2^2 D_{22}}{\rho_2^2} + \frac{2p_1p_2 D_{12}}{\rho_1\rho_2},\\
    \check A &:= \phi \ntrace \Theta K_0^2 K^{-2} = \frac{p_1^2 \check D_{11}}{\rho_1^2} + \frac{p_2^2 \check D_{22}}{\rho_2^2} + \frac{2p_1p_2 \check D_{12}}{\rho_1\rho_2},\\
    \alpha &:= \frac{\phi^2 E_1/\rho_1^2}{1-\psi A},\quad \beta := \frac{\phi^2 E_2/\rho_2^2}{1-\psi A},\quad \tau := \frac{\phi \check A}{1-\psi A},\\
\tilde z_0 &:= \phi \tilde F_0 + \psi\tau \tilde E_0,\quad z_i := \phi F_i + \psi \tau E_i\text{ for }i \in \{1,2\},\\
u &:= \psi D_{11}/\rho_1^2 + \alpha E_1,\quad v:=\psi D_{12}/\rho_2^2 + \beta E_1,\quad w := \psi D_{12}/\rho_1^2 + \alpha E_2,\\
r &:= \psi D_{22}/\rho_2^2 + \beta E_2,\quad
x := \psi \tilde E_1/\rho_1^2 + \alpha \tilde E_0,\quad y := \psi \tilde E_2/\rho_2^2 + \beta \tilde E_0,
\end{split}
}
\label{eq:scalars}
\end{eqnarray}
where $\Sigma_0 := I_d$. Here, the $\Sigma_k$ are the covariance matrices for the different groups, and the proportions $p_1,p_2,\phi,\psi$ are as in the asymptotic limits \eqref{eq:proportionate}.

\subsection{Analysis of Variance}
WLOG, set $\sigma_2=0$. We only need show that
$$
Var \simeq \sigma_1^2 \frac{\det U^{(1)}_{1 \leftarrow V^{(1)}}}{\det U^{(1)}}.
$$

\subsubsection{Linear Pencil Construction}
Consider $X_1 W' H W B W' H W X_1'  \to r:=z_1 s_1 w' h w b w' h w s_1 z_1' $, with $h=(t + \lambda)^{-1}$, $t=t_1 + t_2$, $t_k = ws_k z_k' z_k s_k w'$, and the mapping
\begin{align}
    s_k &\to S_k := \Sigma_k^{1/2},\quad
    b \to B,\quad
    z_k \to Z_k := X_k \Sigma_k^{-1/2}/\sqrt{n}.
\end{align}
Note that that $Z_k$'s are independent with iid entries from $\mathcal N(0,1/n)$.

Now, let $r_1=w'hw s_1 z_1'$ ($d \times n_k$), so that $r=r_1' b r_1$ ($n_k \times n_k$). For input $u_0=1$, the output is $u_1 = r_1 u_0 = w'u_2$, $u_2=hw s_1 z_1' u_0$. The equation for $u_2$ is equivalent to
$$
ws_1z_1' u_0 = (\lambda+t)u_2 = \lambda u_2 + t_1 u_2 + t_2 u_2.
$$
Now, $t_1 u_2 = ws_1 z_1' z_1 s_1 w' u_2 = wu_3$, $u_3=s_1 u_4$, $u_4=z_1'u_5$, $u_5=z_1 u_6$, $u_6=s_1 u_1$. Likewise, $t_2 u_2 = w s_2 z_2' z_2 s_2 w' u_2 = wu_7$, $u_7 = s_2 u_8$, $u_8 = z_2' u_9$, $u_9 = z_2 u_{10}$, $u_{10} = s_2 u_1$.

Finally, $w s_1 z_1' u_0 = w u_{11}$, $u_{11}= s_1 u_{12}$, $u_{12} = z'_1 u_0$. We get following linear system
\begin{align*}
    u_0 &= 1\\
    u_1 - w' u_2 &= 0\\
    \lambda u_2 + w u_3 + w u_7 - w u_{11} &= 0\\
    u_3 - s_1 u_4 &= 0\\
    u_4 - z_1' u_5 &= 0\\
    u_5 - z_1 u_6 &= 0\\
    -s_1 u_1 + u_6 &= 0\\
    u_7 - s_2 u_8 &= 0\\
    u_8 - z_2' u_9 &= 0\\
    u_9 - z_2 u_{10} &= 0\\
    -s_2 u_1 + u_{10} &= 0\\
    u_{11} - s_1 u_{12} &= 0\\
    -z_1' u_0 + u_{12} &= 0
\end{align*}
That is $13$ equations in $13$ unknowns $u=(u_0,u_1,\ldots,u_{12})$, which can be packaged as $Q_1u=e_1$, with
\begin{eqnarray*}
\begin{split}
    Q_1 &= \begin{pmatrix}
    1 & 0 & 0 & 0 & 0 & 0 & 0 & 0 & 0 & 0 & 0 & 0 & 0\\
    0 & 1 & -w' & 0 & 0 & 0 & 0 & 0 & 0 & 0 & 0 & 0 & 0\\
    0 & 0 & 0 & 1 & -s_1 & 0 & 0 & 0 & 0 & 0 & 0 & 0 & 0\\
    0 & 0 & 0 & 0 & 1 & -z_1' & 0 & 0 & 0 & 0 & 0 & 0 & 0\\
    0 & 0 & 0 & 0 & 0 & 1 & -z_1 & 0 & 0 & 0 & 0 & 0 & 0\\
    0 & -s_1 & 0 & 0 & 0 & 0 & 1 & 0 & 0 & 0 & 0 & 0 & 0\\
    0 & 0 & 0 & 0 & 0 & 0 & 0 & 1 & -s_2 & 0 & 0 & 0 & 0\\
    0 & 0 & 0 & 0 & 0 & 0 & 0 & 0 & 1 & -z_2' & 0 & 0 & 0\\
    0 & 0 & 0 & 0 & 0 & 0 & 0 & 0 & 0 & 1 & -z_2 & 0 & 0\\
    0 & 0 & \lambda & w & 0 & 0 & 0 & w & 0 & 0 & 0 & -w & 0\\
    0 & -s_2 & 0 & 0 & 0 & 0 & 0 & 0 & 0 & 0 & 1 & 0 & 0\\
    0 & 0 & 0 & 0 & 0 & 0 & 0 & 0 & 0 & 0 & 0 & 1 & -s_1\\
    -z_1' & 0 & 0 & 0 & 0 & 0 & 0 & 0 & 0 & 0 & 0 & 0 & 1
\end{pmatrix}
\end{split}
\label{eq:Q1}
\end{eqnarray*}
This gives a linear pencil for $r_1$ via $r_1=u_1 = e_2' u = e_2' Q_1^{-1} e_1 = Q_1^{-1}[2,1]$. A linear pencil for $r=r_1'br_1$ is then the $26 \times 26$ matrix $Q$ given by
\begin{eqnarray}
    Q = \begin{pmatrix}
        0 & Q_1\\
        Q_1' & -N
    \end{pmatrix},\text{ with }N = e_2 b e_2' = \begin{pmatrix}
        0 & 0 & 0 & \ldots & 0\\
        0 & b & 0 & \ldots & 0\\
        \vdots & \vdots & \vdots & \ldots & \vdots\\
        0 & 0 & 0 & \ldots &  0
    \end{pmatrix}
\end{eqnarray}
via $r = (Q^{-1})_{1,1}$. In terms of matrices, this pencil has the folliwing block structure
\begin{eqnarray}
    \begin{split}
Q &= \begin{pmatrix}
        0 & Q_1\\
        Q_1' & -N
    \end{pmatrix},\\
    \text{with }N &= \begin{pmatrix}
        0 & 0 & 0 & \ldots & 0\\
        0 & B & 0 & \ldots & 0\\
        \vdots & \vdots & \vdots & \ldots & \vdots\\
        0 & 0 & 0 & \ldots &  0
    \end{pmatrix},\\
    Q_1 &= \begin{pmatrix}
    I_{n_1} & 0 & 0 & 0 & 0 & 0 & 0 & 0 & 0 & 0 & 0 & 0 & 0\\
    0 & I_d & -W' & 0 & 0 & 0 & 0 & 0 & 0 & 0 & 0 & 0 & 0\\
    0 & 0 & 0 & I_d & -\Sigma_1^{1/2} & 0 & 0 & 0 & 0 & 0 & 0 & 0 & 0\\
    0 & 0 & 0 & 0 & I_d & -Z_1' & 0 & 0 & 0 & 0 & 0 & 0 & 0\\
    0 & 0 & 0 & 0 & 0 & I_{n_1} & -Z_1 & 0 & 0 & 0 & 0 & 0 & 0\\
    0 & -\Sigma_1^{1/2} & 0 & 0 & 0 & 0 & I_d & 0 & 0 & 0 & 0 & 0 & 0\\
    0 & 0 & 0 & 0 & 0 & 0 & 0 & I_d & -\Sigma_2^{1/2} & 0 & 0 & 0 & 0\\
    0 & 0 & 0 & 0 & 0 & 0 & 0 & 0 & I_d & -Z_2' & 0 & 0 & 0\\
    0 & 0 & 0 & 0 & 0 & 0 & 0 & 0 & 0 & I_{n_2} & -Z_2 & 0 & 0\\
    0 & 0 & \lambda I_m & W & 0 & 0 & 0 & w & 0 & 0 & 0 & -W & 0\\
    0 & -\Sigma_2^{1/2} & 0 & 0 & 0 & 0 & 0 & 0 & 0 & 0 & I_d & 0 & 0\\
    0 & 0 & 0 & 0 & 0 & 0 & 0 & 0 & 0 & 0 & 0 & I_d & -\Sigma_1^{1/2}\\
    -Z_1' & 0 & 0 & 0 & 0 & 0 & 0 & 0 & 0 & 0 & 0 & 0 & I_d
\end{pmatrix},
    \end{split}
\end{eqnarray}
and $X_1 W' H W B W' H W X_1' = Q^{-1}[1,1]$, and so
$$
\mathbb E[\frac{1}{n}\trace X_1 W' H W B W' H W X_1'] = \frac{n_1}{n}\mathbb E\ntrace Q^{-1}[1,1] \simeq p_1 \mathbb E\ntrace Q^{-1}[1,1].
$$

\subsubsection{Fixed-Point Equations}
Henceforth we assume that $\Sigma_1$, $\Sigma_2$, and $B$ commute. Using auto-fpt gives $V_1 = G_{0,0}$, with
\begin{align*}
    G_{0, 0} &= \frac{\phi G_{3,3}}{(\phi G_{3,19} - 1)^2},\\
    G_{3,3} &= \psi \ntrace \Sigma_1 K^{-2}L,\\
    K &= \phi I_d + \psi G_{9,15}\bar K,\quad \bar K = p_1G_{4,18}\Sigma_1 + p_2 G_{8,22}\Sigma_2,\\
    L &= \phi G_{9,9}I_d + \psi G_{9,15}^2\left(M + B\right),\quad M=p_1 G_{4,4}\Sigma_1 + p_2 G_{8,8}\Sigma_2,\\
    G_{4,4} &= G_{0,0},\\
    G_{8,8} &= \frac{\phi G_{7,7}}{(\phi G_{7,23} - 1)^2},\\
    G_{7,7} &= \psi\ntrace\Sigma_2 K^{-2} L,\\
    G_{3,19} &= -\psi G_{9,15}\ntrace\Sigma_1 K^{-1},\\
    G_{7,23} &= -\psi G_{9,15}\ntrace\Sigma_2 K^{-1},\\
    G_{9,15} &= \frac{1}{\lambda + G_{1,16} + G_{1,20}},\\
    G_{1,16} &= p_1 G_{4,18}\phi\ntrace\Sigma_1 K^{-1},\\
    G_{1,20} &= p_2 G_{8,22}\phi\ntrace\Sigma_2 K^{-1},\\
    G_{4,18} &= \frac{1}{1-\phi G_{3,19}},\\
    G_{8,22} &= \frac{1}{1-\phi G_{7,23}},\\
    G_{9,9} &= G_{1,1}G_{9,15}^2,\\
    G_{1,1} &= \phi\ntrace K^{-2} (\phi C + \psi G_{9,9}\bar K^2 + \phi M).
\end{align*}
Define $1/\delta = G_{9,15}$, $1/\rho_1 = G_{4,18}/\delta$, and $1/\rho_2 = G_{8,22}/\delta$. This gives
$$
K = \phi I_d + \frac{p_1\psi}{\rho_1}\Sigma_1 + \frac{p_2\psi}{\rho_2}\Sigma_2.
$$
Equation for $\delta$ gives $\delta = \lambda  + G_{1,16} + G_{1,20}$. But
\begin{align*}
    G_{1,16} &= p_1 G_{4,18}\phi \ntrace\Sigma_1 K^{-1} = \frac{p_1\delta}{\rho_1}\phi \ntrace\Sigma_1 K^{-1},\\
    G_{1,20} &= p_2 G_{8,22}\phi\ntrace\Sigma_2 K^{-1} = \frac{p_1\delta}{\rho_1}\phi \ntrace\Sigma_1 K^{-1}.
\end{align*}
Now, observe that
\begin{align*}
\rho_1/\delta &= 1/G_{4,18} = 1 + \phi \psi G_{9,15}\ntrace\Sigma_1 K^{-1} = 1 + (\phi\ntrace \Sigma_1 K^{-1})\psi /\delta,\\
\rho_2/\delta &= 1/G_{8,22} = 1 + \phi \psi G_{9,15}\ntrace\Sigma_2 K^{-1}= 1 + (\phi \ntrace \Sigma_2 K^{-1})\psi /\delta,
\end{align*}
and so $\phi \ntrace \Sigma_j K^{-1} = (\rho_j - \delta)/\psi$, that is
\begin{eqnarray}
    \rho_j = \delta + \psi \mu_j,\text{ with }\mu_j := \phi\ntrace\Sigma_jK^{-1}. 
\end{eqnarray}
We deduce that
\begin{align*}
    G_{1,16} &= \frac{p_1\delta \mu_1}{\rho_1} = \frac{p_1\delta }{\psi}(1 - \frac{\delta}{\rho_1}),\\
    G_{1,20} &= \frac{p_2\delta \mu_2}{\rho_2} = \frac{p_2\delta }{\psi}(1 - \frac{\delta}{\rho_2}),\\
\end{align*}
Therefore the fixed-point equations for $(\delta,\rho_1,\rho_2)$ are
\begin{eqnarray*}
    \begin{split}
    \psi \lambda &= (\psi - 1)\delta + \delta^2\left(\frac{p_1}{\rho_1} + \frac{p_2}{\rho_2}\right),\\
    \text{with }\rho_j &:= \delta + \psi\mu_j,\quad \mu_j := \phi\ntrace\Sigma_j K^{-1},\quad K := \phi I_d + \frac{p_1\psi}{\rho_1}\Sigma_1 + \frac{p_2\psi}{\rho_2}\Sigma_2.
    \end{split}
\end{eqnarray*}
We then get
\begin{align*}
    L &= \frac{1}{\delta^2}\left(\psi M + \psi B + \phi G_{1,1} I_d\right),\\
    G_{0,0} &= \frac{\psi\nu_1}{(1+\psi \mu_1/\delta)^2} = \frac{\psi \nu_1\delta^2}{\rho_1^2},\\
    \nu_1 &= \phi\ntrace\Sigma_1 K^{-2}L=\frac{1}{\delta^2} \phi \ntrace \Sigma_1 K^{-2}(p_1 \psi G_{0,0}\Sigma_1 + p_2 \psi G_{8,8}\Sigma_2 + \psi B + \phi G_{1,1}I_d)\\
    &= \frac{p_1 \psi D_{11}G_{0,0} + p_2 \psi D_{12}G_{8,8} + \phi E_1 G_{1,1} + \psi \tilde E_1}{\delta^2},\\
 D_{ij} &:= \phi \ntrace \Sigma_i\Sigma_j K^{-2},\quad \tilde E_i := \phi \ntrace B\Sigma_i  K^{-2}, \quad E_i := \phi\ntrace\Sigma_j K^{-2}, \\
    G_{8,8} &= \frac{\psi \nu_2}{(1+ \mu_2/\delta)^2},\\
\nu_2 &= \phi \ntrace\Sigma_2 K^{-2} L = \ldots= \frac{p_1\psi D_{12}G_{0,0} + p_2 \psi D_{22}G_{8,8} + \phi E_2 G_{1,1} + \psi \tilde E_2}{\delta^2},\\
G_{1,1} &= \phi \ntrace K^{-2}(\phi B + \psi (G_{1,1}/\delta^2)\bar K^2 + \phi M) 
= \psi AG_{1,1} + \phi\cdot (p_1 E_1 G_{0,0} + p_2 E_2 G_{8,8} + \tilde E_0),\\
 \tilde F_0 &:= \phi \ntrace BK^{-2},\quad A := \phi\ntrace K^{-2}\left(\frac{p_1}{\rho_1}\Sigma_1  + \frac{p_2}{\rho_2}\Sigma_2\right)^2 = \frac{p_1^2 D_{11}}{\rho_1^2} + \frac{p_2^2 D_{22}}{\rho_2^2} + \frac{2p_1p_2 D_{12}}{\rho_1\rho_2}.
\end{align*}

\subsubsection{Analytic Formula}
\label{sec:var-analytic}
Rearranging the above equations gives the following linear system of 3 equations in 3 unknowns $(G_{0,0},G_{8,8}, G_{1,1})$:
\begin{align*}
    (\rho_1^2/\psi^2 - p_1 D_{11})G_{0,0} - p_2 D_{12}G_{8,8} -  (E_1/\gamma) G_{1,1} &= \tilde E_1,\\
    -p_1 D_{12}G_{0,0} + (\rho_2^2/\psi^2 - p_2 D_{22})G_{8,8} - (E_2/\gamma) G_{1,2} &=  \tilde E_2,\\
    -p_1 \phi  E_1 G_{0,0} - p_2 \phi  E_2 G_{8,8} + (1 - \psi A)G_{1,1} &= \phi \tilde E_0.
\end{align*}
Therefore by Cramer's rule,
\begin{eqnarray*}
    \begin{split}
            Var \simeq p_1\sigma_1^2 G_{0,0} &= p_1\sigma_1^2 \frac{N}{\Delta},\\
    \text{where }N &:= \det U_{1 \leftarrow V},\quad
    \Delta := \det U,\quad V := \begin{pmatrix}
        \tilde E_1\\\tilde E_2\\\phi \tilde E_0
    \end{pmatrix},\quad U = \begin{pmatrix}
        a_1 & -p_2 D_{12} & - E_1/\gamma\\
        -p_1 D_{12} & a_2 & -E_2/\gamma\\
        -p_1 \phi  E_1 & -p_2 \phi  E_2 & 1 - \psi A
    \end{pmatrix},
    \end{split}
\end{eqnarray*}
where the nonnegative scalars $a_1$, $a_2$, $A$, $D_{12}$, $D_{21}$, $E_1$, $E_2$, $\tilde E_0$, $\tilde E_1$, $\tilde E_2$$u$, $v$, $w$, $r$, $z_1$, $z_2$, given explicitly in \eqref{eq:scalars}. The claimed formula for the bias term then follows.

\subsection{Analysis of Bias}
\subsubsection{Linear Pencil Construction}
Consider the rational expression $w_*' r_1' pp r_1 w_*$ with $r_1=(t+\lambda)^{-1}$ and $t=t_1 + t_2$, $t_j=w'ws_jz_j'z_j s_j$, which is the expression appearing in the bias term, once we make the identification
$$
p \to B^{1/2},\quad w \to W,\quad s_j \to \Sigma_j^{1/2},\quad z_j \to Z_j := X_j\Sigma^{-1/2}/n.
$$
The aforementioned expression has the same trace as $r:=pr_1 a r_1' p$, with $a=w_*w_*'$. If $Q_1$ is a linear pencil for $r_1$ via $r_1=c_1' Q_1^{-1} b_1$, then the block matrix
\begin{eqnarray}
    Q = \begin{pmatrix}
    0 & 1 & 0 & 0\\
    1 & 0 & G' & 0\\
    0 & G & 0 & Q_1'\\
    0 & 0 & Q_1 & -H
    \end{pmatrix},\text{ with }G = c_1 p,\quad H=b_1 a b_1'
\end{eqnarray}
is a linear pencil for $r$ via $r=e_1'Q^{-1} e_1$. It is easy to see that
$$
\quad G = \begin{pmatrix}
    p\\0\\\vdots\\0
\end{pmatrix},\quad G' = \begin{pmatrix}
    p & 0 & \ldots & 0
\end{pmatrix},\quad H = \begin{pmatrix}
    a & 0 & \ldots & 0\\
    0 & 0 & \ldots & 0\\
    \vdots & \vdots & \ldots & \vdots\\
    0 & 0 & 0 \ldots & 0
\end{pmatrix}.
$$

Furthermore, a linear pencil for $r_1$ is given by
$$
Q_1 =
\begin{pmatrix}
\lambda & 0 & 0 & 0 & 0 & w' & 0 & 0 & 0 & 0 & w'\\
-s_1 & 1 & 0 & 0 & 0 & 0 & 0 & 0 & 0 & 0 & 0\\
0 & -z_1 & 1 & 0 & 0 & 0 & 0 & 0 & 0 & 0 & 0\\
0 & 0 & -z_1' & 1 & 0 & 0 & 0 & 0 & 0 & 0 & 0\\
0 & 0 & 0 & -s_1 & 1 & 0 & 0 & 0 & 0 & 0 & 0\\
0 & 0 & 0 & 0 & -w & 1 & 0 & 0 & 0 & 0 & 0\\
-s_2 & 0 & 0 & 0 & 0 & 0 & 1 & 0 & 0 & 0 & 0\\
0 & 0 & 0 & 0 & 0 & 0 & -z_2 & 1 & 0 & 0 & 0\\
0 & 0 & 0 & 0 & 0 & 0 & 0 & -z_2' & 1 & 0 & 0\\
0 & 0 & 0 & 0 & 0 & 0 & 0 & 0 & -s_2 & 1 & 0\\
0 & 0 & 0 & 0 & 0 & 0 & 0 & 0 & 0 & -w & 1
\end{pmatrix},
$$
via $r_1=e_1' Q_1^{-1} e_1$.

\subsubsection{Fixed-Point Equations}
\begin{align*}
    G_{0,0} &= \ntrace B H^{-2}L,\\
    H &= p_1 xG_{4,15}\Sigma_1 + p_2 y G_{9,20}\Sigma_2 + \lambda\phi I_d,\quad x=(G_{7,18} + G_{12,18})\psi,\quad y = (G_{7,23} + G_{12,23})\psi,\\
    L &= p_1 G_{4,4}x^2\Sigma_1 + p_2 G_{9,9}y^2 \Sigma_2 + \phi^2\Theta + \phi\psi s I_d,\quad s = G_{7,7} + 2G_{7,12} + G_{12,12},\\
    G_{4,15} &= \frac{1}{1-\phi G_{3,16}},\quad G_{9,20} = \frac{1}{1-\phi G_{8,21}},\quad 
    G_{3,16} = -x \ntrace\Sigma_1 H^{-1}, \quad G_{8,21} =-y\ntrace\Sigma_2 H^{-1}, \\
     G_{7,18} &= \frac{G_{11,13} + 1}{z},\quad G_{12,18} = - \frac{G_{11,13}}{z},\quad z = G_{6,13} + G_{11,13} + 1,\\
    G_{7,23} &= -\frac{G_{6,13}}{z}=G_{7,18} - 1,\quad G_{12,23} = \frac{G_{6,13} + 1}{z} = 1 + G_{12,18},\\
    G_{6,13} &= p_1 G_{4,15}\phi\ntrace \Sigma_1 H^{-1},\quad G_{11,13} = p_2 G_{9,20}\phi\ntrace \Sigma_2 H^{-1},\\
    G_{4,4} &= \phi G_{3,3}G_{4,15}^2,\quad G_{9,9} = \phi G_{8,8}G_{9,20}^2,\quad G_{3,3} = \ntrace \Sigma_1 H^{-2}L,\quad G_{8,8} = \ntrace\Sigma_2 H^{-2} L.
\end{align*}
One easily deduces that
\begin{align*}
     H &= \lambda K,\quad K = \psi K_0 + \phi I_d,\quad K_0 = \frac{1}{\lambda z}\left(p_1G_{4,15}\Sigma_1 + p_2G_{9,20}\Sigma_2\right) = \frac{p_1}{\rho_1}\Sigma_1 + \frac{p_2}{\rho_2} \Sigma_2,\\
    G_{4,15} &= \frac{\lambda z}{\rho_1},\quad G_{9,20} = \frac{\lambda z}{\rho_2},\quad
x = y = \frac{\psi}{z},\\
z &= 1+\phi\ntrace(p_1G_{4,15}\Sigma_1 + p_2 G_{9,20}\Sigma_2)H^{-1}\\
&= 1+\phi\ntrace(p_1\frac{G_{4,15}}{\lambda}\Sigma_1 + p_2 \frac{G_{9,20}}{\lambda}\Sigma_2)K^{-1} = 1 + tz,\\
\text{i.e., }z &= \frac{1}{1-t},\quad t := \frac{p_1\mu_1}{\rho_1} + \frac{p_2\mu_2}{\rho_2},\quad
\mu_j = \phi\ntrace\Sigma_j K^{-1},
\end{align*}
where $\delta$, $\rho_1$, $\rho_2$ are as introduced before in \eqref{eq:delta}.
We deduce that $\lambda z = \delta$, and so
\begin{align*}
    \lambda z &=\delta,\quad G_{4,15} = \frac{\delta}{\rho_1},\quad G_{9,20} = \frac{\delta}{\rho_2},\quad x=y=\frac{\psi\lambda}{\delta},\\
    s &= \frac{(\phi^2/z^2)\ntrace M K^{-2} + \phi A}{1-\psi A},\quad M=p_1 G_{4,4}\Sigma_1 + p_2 G_{9,9}\Sigma_2,\\
    c&:=\lambda^2 G_{0,0} = \ntrace B K^{-2} L = \ntrace B K^{-2}\left(\phi a\frac{p_1 \psi^2}{\rho_1^2} \Sigma_1 + \phi b \frac{p_2\psi^2}{\rho_2^2} \Sigma_2 + \phi^2\Theta + \phi\psi s I_d\right)\\
    &= \frac{p_1 \psi^2 \tilde E_1}{\rho_1^2}a + \frac{p_2 \psi^2 \tilde E_2}{\rho_2^2}b + \phi \tilde F_0 + \psi s \tilde E_0,\\
    a &:= \lambda^2 G_{3,3},\quad b := \lambda^2 G_{8,8},\\ \tilde E_0 &:= \phi\ntrace B K^{-2},\quad \tilde F_0 := \phi\ntrace \Theta B K^{-2},\quad \tilde E_1 := \phi \ntrace B \Sigma_1 K^{-2},\quad \tilde E_2 := \phi\ntrace B\Sigma_2 K^{-2}.
\end{align*}
Observe that we must have
\begin{align*}
        s &= \frac{(p_1 E_1 a/\rho_1^2 + p_2 E_2 b/\rho_2^2)\phi^2 + \phi \check A}{1-\psi A},\quad E_j := \phi\ntrace \Sigma_j K^{-2},\\
        A &:= \phi\ntrace \Sigma^2 K^{-2},\quad \check A := \phi \ntrace\Theta K_0^2 K^{-2},\\
        a &= \ntrace \Sigma_1 K^{-2}L = \frac{p_1 \psi^2 D_{11}}{\rho_1^2}a + \frac{p_2\psi^2 D_{12}}{\rho_2^2}b + \phi F_1 + \psi s E_1,\\
        b &= \ntrace \Sigma_2 K^{-2}L = \frac{p_1\psi^2 D_{12}}{\rho_1^2}a + \frac{p_2 \psi^2 D_{22}}{\rho_2^2}b + \phi  F_2 + \psi s E_2,\\
        c &= \frac{p_1\psi^2 \tilde E_1}{\rho_1^2}a + \frac{p_2\psi^2 \tilde E_2}{\rho_2^2}b + \phi \tilde F_0 + \psi s \tilde E_0.
\end{align*}
It is easy to see that $s=\alpha a + \beta b + \tau$, where
\begin{align*}
     \alpha &:= \frac{\phi^2 E_1/\rho_1^2}{1-\psi A},\quad \beta := \frac{\phi^2 E_2/\rho_2^2}{1-\psi A},\quad \tau := \frac{\phi \check A}{1-\psi A}.   
\end{align*}
Eliminating $s$ from the above equations for $a$, $b$, and $c$, we obtain the following system of linear equations:
\begin{align*}
    &(1 - p_1 \psi u)a - p_2 \psi v b = \phi F_1 + \psi \tau E_1,\\
    &-p_1\psi w a + (1 - p_2\psi r)b = \phi F_2 + \psi \tau E_2,\\
    &c = p_1\psi (\psi \tilde E_1/\rho_1^2  + \alpha\tilde E_0)a +  p_2\psi (\psi \tilde E_2/\rho_2^2 + \beta \tilde E_0)b + \phi \tilde F_0 + \psi\tau \tilde E_0,\\
    &\text{where }u := \psi D_{11}/\rho_1^2 + \alpha E_1,\quad v:=\psi D_{12}/\rho_2^2 + \beta E_1,\quad w := \psi D_{12}/\rho_1^2 + \alpha E_2,\\
    r &:= \psi D_{22}/\rho_2^2 + \beta E_2.
\end{align*}
Solving the first two equations for $a$ and $b$ and plugging into the third gives the following analytic formula for the bias term in the model's test error:
\begin{eqnarray*}
    \begin{split}
        Bias \simeq \lambda^2 G_{0,0} = c &= p_1\psi (\psi \tilde E_1/\rho_1^2  + \alpha\tilde E_0)a +  p_2\psi (\psi \tilde E_2/\rho_2^2 + \beta\tilde E_0)b + \phi \tilde F_0 + \psi\tau \tilde E_0,\\
        &= p_1\psi x a + p_2\psi y b + z_0,\\
        \text{where }\begin{pmatrix}
            a\\b
        \end{pmatrix} &= \begin{pmatrix}
            1-p_1\psi u \quad& \quad-p_2\psi v\\
            -p_1\psi w \quad& \quad1-p_2\psi r
        \end{pmatrix}^{-1}\begin{pmatrix}
            \phi F_1 + \psi \tau E_1\\
            \phi  F_2 + \psi \tau E_2
        \end{pmatrix},
    \end{split}
\end{eqnarray*}
as claimed. This completes the proof of Theorem \ref{thm:main}. \qed

\section{Case of Single Dataset (Proof of Corollary \ref{corr:single})}
\label{sec:single}
\subsection{Variance Term}
As a sanity check, let us consider the limiting case of a single dataset, i.e $p_2 \to 0$ (i.e., $p_1 \to 1$). Setting $B=\Sigma_1=\Sigma$, the fixed-point equations \eqref{eq:delta} become:
\begin{align}
    \psi \lambda &= (\psi - 1 + \frac{\delta}{\rho_1})\delta,\quad
    \rho_1 = \delta + \psi \phi \ntrace\Sigma K^{-1},\quad
    K = \phi I_d + \frac{\psi}{\rho_1}\Sigma.
\end{align}
Setting $\kappa := \phi \rho_1/\psi = \rho_1/\gamma$ and $I = \ntrace \Sigma(\Sigma + \kappa I_d)^{-1}$, we get
$$
\phi \ntrace\Sigma K^{-1} = \kappa I,\quad \delta = \rho_1 - \psi \phi\ntrace\Sigma K^{-1} = \gamma \kappa - \psi \kappa I = (1 - \phi I)\gamma \kappa,\quad \delta/\rho_1 = 1 - \phi I.
$$
Fixed-point equation becomes $\psi \lambda = (\psi - 1 + 1 - \phi I)(1 - \phi I)\kappa = \psi \kappa (\gamma -  I)(1-\phi I)$. Dividing through by $\psi$, this can be written as 
\begin{eqnarray}
    \lambda = (\gamma - I)(1 - \phi I)\kappa,
    \label{eq:kappa}
\end{eqnarray}
an equation encountered a previous lecture. Furthermore, in this case we have
\begin{eqnarray}
\begin{split}
        G_{0,0} &= \frac{\det\begin{pmatrix}
        F_1 & 0 & -E_1/\gamma\\
        F_2 & a_2 & -E_2/\gamma\\
        \phi F_0 & 0 & b
    \end{pmatrix}}{\det \begin{pmatrix}
        a_1 & 0 & -E_1/\gamma\\
        -D_{12} & a_2 & -E_2/\gamma\\
        -\phi E_1 & 0 & b
        \end{pmatrix}}\\
        &= \frac{bF_1 + (\phi/\gamma) E_1 F_0}{a_1 b - (\phi/\gamma) E_1^2} = \frac{(1-\psi A)F_1 + (\phi/\gamma) E_1 F_0}{(\rho^2/\psi^2 - D_{11})(1-\psi A) - (\phi/\gamma) E_1^2}.
\end{split}
\end{eqnarray}
For brevity, write $\bar\df_i := \bar\df_i(\Sigma_1;\kappa) := \ntrace\Sigma^i(\Sigma+\kappa I_d)^{-i}$. Observe that
\begin{align*}
\rho_1^2/\psi^2 &= \kappa^2 \gamma^2/\psi^2 = \kappa^2/\phi^2,\\
D_{11} &= \phi \ntrace\Sigma_1^2(\phi I_d + (\psi/\rho_1)\Sigma_1)^{-2} = \phi \frac{\rho_1^2}{\psi^2}\ntrace\Sigma_1^2(\Sigma_1 + \kappa I_d)^{-2} =  \frac{\kappa^2}{\phi} \bar \df_2,\\
\psi A  &=  \frac{\psi }{\rho_1^2}D_{11}= \frac{\psi}{\gamma^2 \kappa^2} \frac{\kappa^2}{\phi}\bar\df_2 = \frac{1}{\gamma}\bar\df_2,\\
E_1 &= \phi\ntrace\Sigma_1(\phi I_d + (\psi/\rho_1)\Sigma_1)^{-2} =   \frac{\kappa^2}{\phi} J,\\    
F_1 &= \phi \ntrace\Sigma_1^2(\phi I_d + (\psi/\rho_1)\Sigma_1)^{-2} =   \frac{\kappa^2}{\phi}\bar \df_2,\\
F_0 &= \phi \ntrace\Sigma_1(\phi I_d + (\psi/\rho_1)\Sigma_1)^{-2} = \frac{\kappa^2}{\phi}J,\\
\frac{\phi}{\gamma}E_1^2 &= \frac{\phi}{\gamma \phi^2} (\kappa J)^2 \kappa^2=  \frac{\kappa^2}{\psi} (\kappa J)^2,\\
\frac{\phi}{\gamma} E_1 F_0 &=  \frac{\phi}{\gamma \phi^2} (\kappa J)^2 \kappa^2=  \frac{\kappa^2}{\psi} (\kappa J)^2.
\end{align*}
\begin{align*}
    G_{0,0} = \frac{(1-\bar\df_2/\gamma) \bar\df_2/\phi + \kappa^2 J^2/\psi}{(1/\phi-\bar \df_2)(1-\bar\df_2/\gamma)/\phi -  \kappa^2 J^2/\psi} =  \frac{(1-\bar\df_2/\gamma) \bar\df_2 + \kappa^2 J^2/\gamma}{(1/\phi-\bar \df_2)(1-\bar\df_2/\gamma) -  \kappa^2 J^2/\gamma}
\end{align*}
Now, using $\bar\df_2 = I - \kappa J$, one computes $\gamma$ times the numerator and denominator respectively as follows:
\begin{align*}
    (\gamma-\bar\df_2) \bar\df_2 + \kappa^2 J^2 &= \gamma \bar\df_2 + (\kappa^2 J^2 - \bar\df_2^2) = \gamma \bar \df_2 + I (\kappa J - \bar \df_2) = (\gamma - I)\bar\df_2 + \kappa JI\\
    &= (\gamma - I)(I - \kappa J) - \kappa J I = (\gamma - I)I - (\gamma - I)\kappa J + \kappa J I\\
    &= (\gamma - I)I+ (2I - \gamma)\kappa J,\\
(1/\phi-\bar \df_2)(\gamma-\bar\df_2) -  \kappa^2 J^2 &= 1/\phi - (\gamma + 1/\phi)\bar\df_2 + \bar\df_2^2 - \kappa^2 J^2\\
&= \gamma/\phi -(\gamma+1/\phi)\bar\df_2 + (\bar\df_2 - \kappa J)I\\
&= \gamma/\phi + (I - \gamma - 1/\phi)\bar\df_2 - \kappa J I \\
&= \gamma/\phi + (I  -  \gamma - 1/\phi)(I - \kappa J) - \kappa J I\\
&= \gamma/\phi + (I - \gamma)(I - \kappa J) - I/\phi + \kappa J/\phi - \kappa JI \\
&= (I - \gamma)(I - \kappa J) + (\gamma - I)/\phi + (1/\phi - I)\kappa J\\
&= (I - \gamma)(I - 1/\phi) - (I - \gamma)\kappa J + (1/\phi - I)\kappa J\\
&= (\gamma - I)(1 - \phi I)/\phi + (1/\phi + \gamma - 2I)\kappa J\\
&= (\gamma - I)(1 - \phi I)/\phi + (1 + \psi - 2\phi I)\kappa J/\phi
\end{align*}
We deduce that the variance term $Var$ in the test error of the model is given by the following formula:
\begin{eqnarray*}
    \frac{Var}{\sigma^2} \simeq G_{0,0} = \phi\frac{(\gamma - I)I+ (2I - \gamma)\kappa J}{(\gamma - I)(1 - \phi I) + (1 + \psi - 2\phi I)\kappa J},
\end{eqnarray*}
as claimed. \qed



\subsection{Bias Term}
We consider the corner case $p_2 \to 0^+$, with $B=\Sigma_1=\Sigma$. This gives the simplification $c=a$, and we get
\begin{align*}
Bias &\simeq a,\text{ with }\\
    a &= e_1'\begin{pmatrix}
    1-\psi u & 0\\
    \psi w & 1
\end{pmatrix}^{-1}\begin{pmatrix}
    \phi \tilde F_1 + \psi \tau E_1\\
    \phi \tilde F_2 + \psi \tau E_2
\end{pmatrix}\\
&= \frac{e_1'}{1-\psi u}\begin{pmatrix}
    1 & 0\\
    -\psi w & 1-\psi u
\end{pmatrix}\begin{pmatrix}
    \phi \tilde F_1 + \psi \tau E_1\\
    \phi \tilde F_2 + \psi \tau E_2
\end{pmatrix} = \frac{\phi \tilde F_1 + \psi \tau E_1}{1-\psi u}.
\end{align*}
Now, setting $\kappa := \rho_1/\gamma$ which is the unique positive solution to the fixed-point equation \eqref{eq:kappa}, we have
\begin{align*}
\psi A &= D/\gamma,\quad \phi^2 E_1^2/\rho_1^2 = \frac{\phi^2}{\kappa^2 \gamma^2}\frac{\kappa^4 J^2}{\phi^2} = \frac{\kappa^2 J^2}{\gamma^2},\\
\alpha E_1 &= \frac{\phi^2 E_1^2/\rho_1^2}{1-\psi A} = \frac{\kappa^2 J^2/\gamma^2}{1-D/\gamma},\text{ with }D:=\ntrace \Sigma^2(\Sigma + \kappa I_d)^{-2},\\
u &=  \psi D_{1,1}/\rho_1^2 + \alpha E_1 = (\psi/\gamma^2) D_{11}/\kappa^2 +  \frac{\kappa^2 J^2/\gamma^2}{1-D/\gamma} = (\psi/\gamma^2)D/\phi = D/\gamma +  \frac{\kappa^2 J^2/\gamma^2}{1-D/\gamma},\\
1-\psi u &= \frac{(1-\phi D)(1-D/\gamma) - \phi \kappa^2 J^2/\gamma}{1-D/\gamma},\\
\phi\check A &= \phi \tilde D/\gamma ,\quad \phi \tilde F_1 + \psi\tau E_1 = \kappa^2 \tilde J + \frac{\phi \tilde D/\gamma}{1-D/\gamma}\frac{\kappa^2 J}{\phi} = \frac{\kappa^2}{1-D/\gamma}\left((1-D/\gamma)\tilde J + J \tilde D/\gamma\right),\\
\text{with }\tilde J &:= \ntrace \Theta \Sigma(\Sigma + \kappa I_d)^{-2},\quad \tilde D := \ntrace\Theta \Sigma^2(\Sigma + \kappa I_d)^{-2}.
\end{align*}
We deduce that in this special case, the bias term of the test error is given by
\begin{eqnarray*}
    Bias \simeq \kappa^2 \frac{(1-D/\gamma)J + J \tilde D/\gamma}{(1-D/\gamma)(1-\phi D) - \phi \kappa^2 J^2/\gamma},
\end{eqnarray*}
as claimed. The treatment of the variance term is analogous. \qed

\section{Multiple-Descent Phenomena}

\subsection{Triple-Descent on Heterogeneous Covariance: Proof of Theorem \ref{thm:triple}, Corollary \ref{corr:Zipf}, and Theorem \ref{thm:K-fold}}
Since $B_2=\varepsilon B_1$, the two curves have identical stationary points and monotonicity intervals. It therefore suffices to analyze $B_1$.

We start with the proof of Theorem \ref{thm:triple}. For simplicity of notation, set $p:=p_1 \in (0,1)$ and $q:=p_2=1-p$. Applying Theorem \ref{thm:main} with the proportional-covariance bias formula gives (after some tedious algebra):
\[
B_1=B_1(\psi,\varepsilon)
=
\frac{\eta(\psi/\phi)}{2(1-\psi)}
\left(
1+\frac{L_\varepsilon(\psi)}{D_\varepsilon(\psi)}
\right),
\quad 0<\psi<1,
\]
where
\begin{align*}
D_\varepsilon(\psi)
&=
\sqrt{
\bigl((1-\varepsilon)\psi-p+q\varepsilon\bigr)^2
+4pq\varepsilon
},\quad 
L_\varepsilon(\psi)
=
p+q\varepsilon+(q-p)(1-\varepsilon)\psi,
\end{align*}
and $\eta(z)>0$ is defined by as in the theorem, namely
\[
\frac1d\sum_{j=1}^d
\frac{s_j}{\eta(z)+zs_j}=1.
\]

For fixed $\psi\ne p$, letting $\varepsilon\to0^+$ gives
\[
b_0(\psi)
:=
\lim_{\varepsilon\to0^+}B_1(\psi,\varepsilon)
=
\begin{cases}
\displaystyle
\eta(\psi/\phi)\frac{p}{p-\psi},
&0<\psi<p,\\[8pt]
\displaystyle
\eta(\psi/\phi)\frac{q\psi}{(\psi-p)(1-\psi)},
&p<\psi<1.
\end{cases}
\]
This proves Theorem \ref{thm:triple}. The proof of Theorem \ref{thm:K-fold} (correspond to $K \ge 3$ groups) is completely analogous and is therefore omitted.

We now turn to the proof of Corollary \ref{corr:Zipf}. First note that the convergence of $B_1$ to $b_0$ above holds together with the first two derivatives on compact subsets of $(0,p)$ and $(p,1)$.
To determine the derivative signs, define
\[
a_j(\psi)
=
\frac{(\psi/\phi)s_j}
{\eta(\psi/\phi)+(\psi/\phi)s_j},
\quad
f_r(\psi)=\frac1d\sum_{j=1}^d a_j(\psi)^r,
\quad
M(\psi)=\frac{\phi f_2(\psi)}{\psi^2}.
\]
The defining equation for $\eta$ implies $f_1=\psi/\phi$.
Implicit differentiation yields
\[
\frac{d}{d\psi}\log\eta(\psi/\phi)
=
-\frac{M(\psi)}{1-\psi M(\psi)}, \quad 
M'(\psi)
=
-\frac{2(f_1f_3-f_2^2)}
{\phi^2f_1^3(f_1-f_2)}
<0.
\]
Indeed, $0<a_j<1$ implies $f_1>f_2$, while Cauchy--Schwarz gives
$f_1f_3>f_2^2$, strictly because the spectrum is non-constant.
Moreover,
\[
M(\psi)<\frac1\psi,
\quad
M(0^+)
=
\frac{\langle s^2\rangle}
{\phi\langle s\rangle^2},
\]
where $\langle s^r\rangle=d^{-1}\sum_{j=1}^d s_j^r$.

For $0<\psi<p$, differentiation of the limiting bias gives
\[
\frac{b_0'(\psi)}{b_0(\psi)}
=
\frac{1-pM(\psi)}
{(p-\psi)(1-\psi M(\psi))}.
\]
The denominator is positive. Suppose that
\begin{equation}
M(0^+)>\frac1p.
\label{eq:power-law-initial-slope-condition}
\end{equation}
The derivative is then initially negative. Since $M$ strictly decreases
and $M(p)<1/p$, there is exactly one point $a\in(0,p)$ satisfying
\[
M(a)=\frac1p.
\]
At this point, the derivative changes from negative to positive.
Thus, $b_0$ has exactly one local minimum before $p$.
Since $M'(a)<0$, this minimum is nondegenerate.

For $p<\psi<1$,
\[
\frac{b_0'(\psi)}{b_0(\psi)}
=
-\frac{M(\psi)}{1-\psi M(\psi)}
+
\frac{\psi^2-p}{\psi(\psi-p)(1-\psi)}.
\]
This expression is negative on $(p,\sqrt p]$.
On $(\sqrt p,1)$, its sign equals that of
\[
R(\psi)-M(\psi),
\quad
R(\psi)
:=
\frac{\psi^2-p}{\psi((1+p)\psi-2p)}.
\]
A direct calculation gives
\[
R'(\psi)
=
\frac{2p(1-\psi)(\psi-p)}
{\psi^2((1+p)\psi-2p)^2}
>0.
\]
Hence $R-M$ strictly increases. Furthermore,
\[
R(\sqrt p)=0<M(\sqrt p),
\quad
R(1)=1>M(1).
\]
There is therefore exactly one point $b\in(\sqrt p,1)$ at which
the derivative changes from negative to positive.
This is a second nondegenerate local minimum.

We next show that, for positive $\varepsilon$, the singularity at $p$
becomes exactly one rounded maximum. Set
\[
A(\psi):=\frac{\eta(\psi/\phi)}{2(1-\psi)},
\quad
r:=\psi-p,
\]
and abbreviate $D=D_\varepsilon(\psi)$ and $L=L_\varepsilon(\psi)$.
Differentiating the finite-$\varepsilon$ formula gives
\[
\frac{D^3}{A}\,\partial_\psi B_1
=
\frac{A'}{A}(D^3+LD^2)
+
2pq(1-\varepsilon)^2
\bigl[(q-p)\varepsilon-(1+\varepsilon)r\bigr].
\]
Indeed,
\[
D^2
=
\bigl((1-\varepsilon)r+(q-p)\varepsilon\bigr)^2
+4pq\varepsilon,\quad
L'D^2-LDD'
=
2pq(1-\varepsilon)^2
\bigl[(q-p)\varepsilon-(1+\varepsilon)r\bigr].
\]

Denote the right-hand side by $N(\psi,\varepsilon)$.
Since $A$ is smooth and positive near $p$,
\[
N(\psi,\varepsilon)
=
-2pq\,r+O(r^2+\varepsilon),
\quad
\partial_\psi N(\psi,\varepsilon)
=
-2pq+O(|r|+\varepsilon).
\]
These estimates hold uniformly in a sufficiently small fixed
neighborhood of $p$. Consequently, for sufficiently small
$\varepsilon$, $N$ is strictly decreasing there, positive at the left
endpoint, and negative at the right endpoint. It has exactly one zero.
Because $D^3/A>0$, this zero is a local maximum of $B_1$.

Moreover, $N(p,\varepsilon)=O(\varepsilon)$, while
$\partial_\psi N$ remains bounded away from zero.
The mean value theorem therefore gives
\[
\psi_{\max}=p+O(\varepsilon).
\]

The two nondegenerate minima of $b_0$ persist for sufficiently small
$\varepsilon$, by convergence of the first two derivatives away from
$p$. The same convergence rules out additional stationary points on
compact sets outside neighborhoods of $a$, $p$, and $b$.
Near $\psi=0$, the derivative remains negative by the strict inequality
\eqref{eq:power-law-initial-slope-condition}.
Near $\psi=1^-$,
\[
B_1(\psi,\varepsilon)
=
\frac{\eta(1/\phi)}{1-\psi}+O(1),
\quad
\partial_\psi B_1(\psi,\varepsilon)
=
\frac{\eta(1/\phi)}{(1-\psi)^2}+O(1),
\]
uniformly for small $\varepsilon$, so the derivative is positive.
Thus, there are exactly two local minima and one intervening local
maximum on $(0,1)$.

For $\psi>1$, the bias formula becomes
\[
B_1(\psi,\varepsilon)
=
\eta(1/\phi)\frac{\psi}{\psi-1},
\]
and therefore
\[
\partial_\psi B_1(\psi,\varepsilon)
=
-\frac{\eta(1/\phi)}{(\psi-1)^2}<0.
\]
This gives the third decreasing portion of the curve.

Finally, specialize
\eqref{eq:power-law-initial-slope-condition}
to the power-law spectrum. Since
\[
\langle s\rangle=\frac{c}{d}H_{d,\alpha},
\quad
\langle s^2\rangle=\frac{c^2}{d}H_{d,2\alpha},
\]
we obtain
\[
M(0^+)
=
\frac{H_{d,2\alpha}\,d}{\phi H_{d,\alpha}^{\,2}}.
\]
Using $n_1=pd/\phi$,
\[
M(0^+)>\frac1p
\quad\Longleftrightarrow\quad
\frac{H_{d,\alpha}^{\,2}}{H_{d,2\alpha}}<n_1.
\]
This is precisely the condition on $n_1$ in the statement of Corollary \ref{corr:Zipf}, and the proof is complete.
\qed
    

\end{document}